\documentclass[twoside]{article}

\newlength\titlebox 
\usepackage[round]{natbib}

\usepackage{lipsum}
\usepackage{algorithm}
\usepackage[noend]{algpseudocode}
\makeatletter
\renewcommand{\ALG@name}{Procedure}
\makeatother

\usepackage{amsmath,amsthm,amssymb,amsfonts,bbm,bm}

\usepackage{tikz}
\usetikzlibrary{calc, positioning, plotmarks, shapes.misc}
\usepackage[shortlabels]{enumitem}
\usepackage{placeins}
\newcommand\numberthis{\addtocounter{equation}{1}\tag{\theequation}}

\newcommand{\ranmat}[1]{\overline{\mathbf{#1}}}
\newcommand{\ranvec}[1]{{\mathbf{#1}}}

\usepackage{graphicx}
\usepackage{subcaption}
\newtheorem{theorem}{Theorem}

\newtheorem{prop}[theorem]{Proposition}
\newtheorem{obs}[theorem]{Observation}
\newtheorem{hyp}[theorem]{Hypothesis}

\usepackage[siunitx]{circuitikz}

\usepackage{subfiles}

\usepackage{ifthen}
\newboolean{figuresbool}
\setboolean{figuresbool}{true}

\begin{document}

%

%

\title{Exchangeability and Kernel Invariance in Trained MLPs}
\author{\textbf{Russell Tsuchida}\footnotemark[1] \and \textbf{Fred (Farbod) Roosta}\footnotemark[2] \footnotemark[3] \and  \textbf{Marcus Gallagher}\footnotemark[1]}
\date{\small School of Information Technology and Electrical Engineering, University of Queensland\footnotemark[1] \vspace*{-0.2em} \\ 
\small School of Mathematics and Physics, University of Queensland\footnotemark[2] \quad  International Computer Science Institute\footnotemark[3]}

\maketitle

\begin{abstract}
In the analysis of machine learning models, it is often convenient to assume that the parameters are IID. This assumption is not satisfied when the parameters are updated through training processes such as SGD. A relaxation of the IID condition is a probabilistic symmetry known as \emph{exchangeability}. We show the sense in which the weights in MLPs are exchangeable. This yields the result that in certain instances, the layer-wise kernel of fully-connected layers remains approximately constant during training. We identify a sharp change in the macroscopic behavior of networks as the covariance between weights changes from zero.
\end{abstract}

\section{Introduction}
Despite the widespread usage of deep learning in applications~\citep{mnih2015human, pmlr-v70-kalchbrenner17a, silver2017mastering, oord2017parallel}, current theoretical understanding of deep networks continues to lag behind the pursued engineering outcomes. Recent theoretical contributions have considered networks in their randomized initial state, or made strong assumptions about the parameters or data during training. 

For example,~\cite{cho2009kernel, daniely2016toward, JMLR:v18:14-546, tsuchida2017invariance} analyze the kernels of neural networks with random IID distributions. Insightful analysis connecting signal propagation in deep networks to chaos have made similar assumptions~\citep{poole2016exponential, raghu2017expressive}. Clearly the assumption of random IID weights is only valid when the network is in its random initial state. Random matrix theory has recently been applied to neural networks in an attempt to understand the empirical spectral distribution (ESD) of the Hessian~\citep{pennington2017geometry} and the Gram matrix~\citep{pennington2017nonlinear}, but these works have made strong assumptions on the weight and data distributions. 

We investigate the probabilistic symmetry known as exchangeability, yielding insights into the behavior of deep networks. We uncover the striking result that the layer-wise kernel of MLPs with ReLU activations trained with many optimizers \emph{remains constant} up to a scaling factor during training when the network inputs satisfy certain conditions. When the inputs do not satisfy these conditions, we are able to bound the absolute difference between layer-wise kernel and the kernel of the network in its random IID state. Empirically, we show that certain optimizers result in looser bounds, i.e. their kernel diverges more from the kernel of the network in its random IID state.

\section{Background}
\subsection{Notation}
\label{sec:notation}
Random variables, vectors and matrices are denoted by upper case, bold upper case, and bold upper case with overline characters, respectively. Parenthesized superscripts index the layer of the network to which an object belongs. The first and second post-subscripts index the rows and columns of a matrix, respectively. When the row of a matrix is extracted through an index, it is treated as a column vector. Pre-subscripts indicate the iteration of an iterative optimizer. Subscripts $R$ on expectations $\mathbb{E}_R$ indicate that the expectation is taken over the probability distribution of $R$.

Consider an MLP with an input layer and $L$ non-input layers. Denote the number of neurons in layer $0\leq l \leq L$ by $n^{(l)}$. Denote an input to the network by $\mathbf{x} \in \mathbb{R}^{n^{(0)}}.$ Denote the $n^{(l-1)} \times n^{(l)}$ random weight matrix connecting layer $l-1$ to layer $l$ by $\ranmat{W}^{(l)}.$ Denote the activation function by $\sigma:\mathbb{R} \to \mathbb{R}.$ We will consider ReLU activations throughout. That is, $\sigma(z)=\text{max}(0, z)$. The $\ell^2$ norm will be denoted $\Vert \cdot \Vert$.

\subsection{Exchangeability}
An \emph{exchangeable} sequence of random variables $(Q_1, Q_2, ...)$ has the property that the joint distribution of the sequence is invariant to finite permutations. That is, a sequence $(Q_i)_{i \geq 1}$ is exchangeable if $(Q_1, Q_2, ...) \stackrel{d}{=} (Q_{\pi(1)}, Q_{\pi(2)}, ...)$ for any finite permutation $\pi$. To aid in readability we will omit the index set in the subscript, so that $(Q_i)_{i \geq 1}$ is the same as $(Q_i)_{i}$.

de Finetti's theorem characterizes infinite exchangeable sequences as mixtures of IID random variables. There are a number of equivalent ways of stating this precisely, one of which is given in Theorem~\ref{thm:aldous1}.
\begin{theorem}\citep{aldous1981representations}
\label{thm:aldous1}
An infinite sequence $\mathbf{Q}=(Q_i)_i$ is exchangeable if and only if there exists a measurable function $f$ such that $(Q_i)_i \stackrel{d}{=} \big( f(A, B_i) \big)_i,$ where $A$ and $\bm{B}$ are mutually IID random variables uniform on $[0, 1]$.
\end{theorem}

Generalizations of Theorem~\ref{thm:aldous1} to multi-dimensional arrays of exchangeable sequences exist~\citep{kallenberg2006probabilistic}. Call a matrix $\ranmat{Q}$ \emph{row and column exchangeable} (RCE) if its joint distribution is invariant to permutations in rows and columns. That is, $\ranmat{Q}$ is RCE if ${(Q_{ji})_{ji}\stackrel{d}{=}(Q_{\pi_1(j)\pi_2(i)})_{ji}}$ for any finite permutations $\pi_1$ and $\pi_2$.
\begin{theorem}\citep{aldous1981representations}
\label{thm:aldous2}
An infinite array ${\ranmat{Q}=(Q_{ji})_{ji}}$ is RCE if and only if there exists a measurable function $f$ such that ${(Q_{ji})_{ji} \stackrel{d}{=} \big( f(A, B_j, C_i, D_{ji}) \big)_{ji},}$ where $A,$ $\ranvec{B},$ $\ranvec{C},$ and $\ranmat{D}$  are mutually IID uniform on $[0, 1]$.
\end{theorem}

Intuition concerning the strength of exchangeability in the broader context of probabilistic symmetries may be aided by the implication graph shown in Figure~\ref{fig:impgraph}. Probabilistic symmetries are discussed at length in Kallenberg's monograph~(\citeyear{kallenberg2006probabilistic}).

\tikzstyle{int}=[draw, rounded rectangle, minimum height=2em, minimum width=8em, align=center]
\tikzstyle{init} = [pin edge={-to,thin,black}]
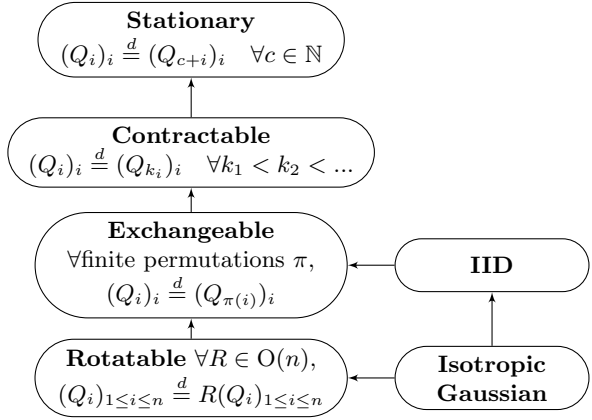
\begin{figure}
\centering
	\begin{tikzpicture}[node distance=0cm,auto,>=latex']
	\tikzstyle{every node}=[font=\small]
    \node [int] (a) 								{\textbf{Stationary} \\$(Q_i)_i \stackrel{d}{=}(Q_{c+i})_i \quad \forall c \in \mathbb{N}$};
    \node [int] (b)	[below of=a,node distance=1.5cm]	{\textbf{Contractable} \\$(Q_i)_i \stackrel{d}{=} (Q_{k_i})_i \quad \forall k_1<k_2<...$};
    \node [int] (c) [below of=b,node distance=1.5cm]	{\textbf{Exchangeable}\\$\forall \text{finite permutations }\pi$,\\$(Q_i)_i \stackrel{d}{=} (Q_{\pi(i)})_i$};
    \node [int] (d)	[below of=c,node distance=1.5cm]	{\textbf{Rotatable} $\forall R \in \text{O}(n),$\\ $(Q_i)_{1 \leq i \leq n} \stackrel{d}{=} R(Q_i)_{1 \leq i \leq n}$};
    \path[->] (b) edge node {} (a);
    \path[->] (c) edge node {} (b);
    \path[->] (d) edge node {} (c);
    
    \node [int] (e)	[right of=d,node distance=4cm]	{\textbf{Isotropic} \\ \textbf{Gaussian}};
    \node [int] (f)	[right of=c,node distance=4cm]	{\textbf{IID}};
    \path[->] (e) edge node {} (f);
    \path[->] (e) edge node {} (d);
    \path[->] (f) edge node {} (c);
\end{tikzpicture}
\caption{Relative strength of probabilistic symmetries found in sequences of random variables.}
\label{fig:impgraph}
\end{figure}

\subsection{Kernels of random MLPs}
\label{sec:randomker}
There is a well-studied connection between the feature maps in MLPs (and other neural network architectures) and the kernel of a reproducing kernel Hilbert Space (RKHS)~\citep{mackay1992practical, neal1995bayesian, cho2009kernel, daniely2016toward, JMLR:v18:14-546, bietti2017invariance}. Consider the angle $\theta^{(l)}$ between two random signals $\bm{\sigma}(\ranmat{W}^{(l)} \mathbf{x})$ and $\bm{\sigma}(\ranmat{W}^{(l)} \mathbf{y})$ in the $l$th hidden layer of an MLP for inputs $\mathbf{x}$ and $\mathbf{y}$. We have
\begin{align*}
\label{eq:finitenormalizedkernel}
&\cos \theta^{(l)} =  \frac{\sum\limits_{j=1}^{n^{(l)}}\sigma \big( \ranvec{W}^{(l)}_j \cdot \mathbf{x} \big) \sigma \big( \ranvec{W}^{(l)}_j \cdot \mathbf{y} \big)}{\sqrt{\sum\limits_{j=1}^{n^{(l)}}\sigma \big( \ranvec{W}^{(l)}_j \cdot \mathbf{x} \big) \sigma \big( \ranvec{W}^{(l)}_j \cdot \mathbf{x} \big)} \sqrt{\sum\limits_{j=1}^{n^{(l)}}\sigma \big( \ranvec{W}^{(l)}_j \cdot \mathbf{y} \big) \sigma \big( \ranvec{W}^{(l)}_j \cdot \mathbf{y} \big)}}, \numberthis
\end{align*}
where $\ranvec{W}^{(l)}_j$ is the $j$th row of $\ranmat{W}^{(l)}$. We can divide the numerator and denominator by $n^{(l)} \Vert \mathbf{x} \Vert \Vert \mathbf{y} \Vert$. Noting the absolute-homogeneity property of the ReLU $\sigma(|a|z) = |a|\sigma(z),$ we consider the scaled numerator
\begin{equation}
\label{eq:finitekernel}
\frac{1}{n^{(l)}}\sum\limits_{j=1}^{n^{(l)}}\sigma \Big( \ranvec{W}_j \cdot \frac{\mathbf{x}}{\Vert \mathbf{x} \Vert} \Big) \sigma \Big( \ranvec{W}_j \cdot \frac{\mathbf{y}}{\Vert \mathbf{y} \Vert} \Big).
\end{equation}
Let $\mathbf{\hat{x}} = \frac{\mathbf{x}}{\Vert \mathbf{x} \Vert}$. Suppose that each row $\ranvec{W}^{(l)}_j$ of $\ranmat{W}^{(l)}$ is IID with other rows (we relax this requirement later) and is defined on some probability space $(\Omega, \Sigma, \mu)$ with $\mu$ not necessarily IID. Asymptotically in the number of neurons $n^{(l)}$, the strong law of large numbers implies that~\eqref{eq:finitekernel} converges almost surely to 
\begin{align*}
\label{eq:infinitekernel}
&\phantom{{}={}}\mathbb{E} \big[ \sigma(\ranvec{W}^{(l)}_j \cdot \mathbf{\hat{x}})  \sigma(\ranvec{W}^{(l)}_j \cdot \mathbf{\hat{y}}) \big] = \int_{\Omega} \sigma( \ranvec{W}^{(l)}_j \cdot \mathbf{\hat{x}}) \sigma( \ranvec{W}^{(l)}_j \cdot \mathbf{\hat{y}}) \,d\mu, \numberthis
\end{align*}
which corresponds to an inner-product in feature space. The kernel is positive semi-definite and uniquely defines an RKHS. When $\mu$ is the product measure corresponding to an IID Gaussian with variance $\mathbb{E}\big[ (W^{(l)}_{11})^2 \big]$ and $0$ mean, the kernel has a closed-form expression known as the arc-cosine kernel (of degree $1$)~\citep{cho2009kernel}, given by
\begin{equation}
\label{eq:arccosker}
\frac{\mathbb{E}\big[ (W^{(l)}_{11})^2 \big]}{2\pi}\big(\sin\theta^{(l-1)} + (\pi-\theta^{(l-1)})\cos\theta^{(l-1)} \big),
\end{equation}
where $\theta^{(l-1)}$ is the angle between $\mathbf{x}$ and $\mathbf{y}$.

We will refer to~\eqref{eq:infinitekernel} as the \emph{layer-wise kernel in layer $l$}, denoted $k^{(l)}(\mathbf{x}, \mathbf{y})$. When~\eqref{eq:infinitekernel} is normalized in the same fashion as~\eqref{eq:finitenormalizedkernel}, we will call the resulting quantity the \emph{layer-wise normalized kernel in layer $l$}.

\subsection{Asymptotic Invariance of Layer-Wise Normalized Kernel in IID Networks}
\label{sec:asymker}
\newcommand\scale{0.2}
\newcommand\step{0.195}
\begin{figure}
\centering
\includegraphics[scale=0.4]{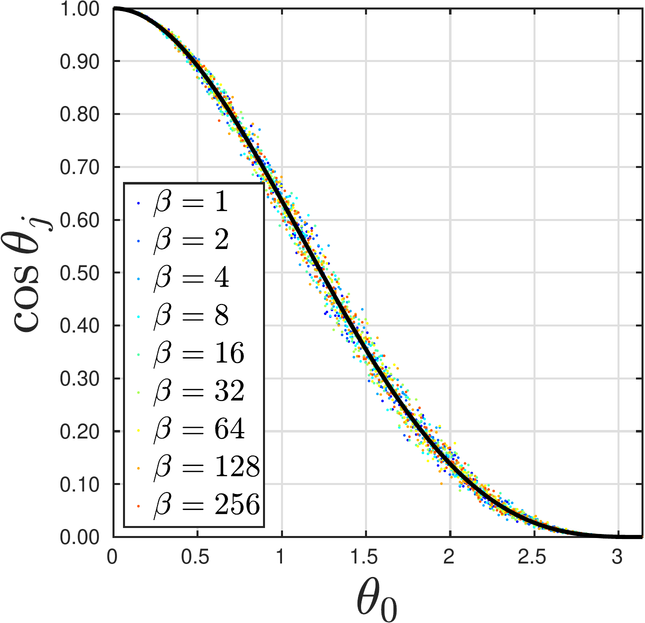}
\caption{Normalized kernel for a hidden layer with ReLU activations. Samples from a network with $1000$ inputs and $1000$ hidden units are obtained by generating $R$ from a $QR$ decomposition of a random matrix containing IID samples from $\mathcal{U}[0,1]$, then setting $\mathbf{x}=R(1, 0)^T$ and $\mathbf{y}=R(\cos\theta, \sin\theta)^T$.}
\label{fig:kernelinit}
\end{figure}
Our analysis draws upon and extends results concerning the layer-wise normalized kernels of MLPs with IID weights~\citep{tsuchida2017invariance}, which, for completeness, we briefly review here.

Construct a sequence $\{ \mathbf{x}_{(m)} \}_{m \geq 2}$ such that for all $m$, $\mathbf{x}_{(m)} \in \mathbb{R}^\infty$ and coordinates $m+1, m+2, ...$ of $\mathbf{x}_{(m)}$ are all $0$. Define the sequence $\{ \mathbf{y}_{(m)} \}_{m \geq 2}$ in the same way, and additionally require that the angle $\theta^{(l-1)}$ between $\mathbf{x}_{(m)}$ and $\mathbf{y}_{(m)}$ is constant in $m$. 

Denote the randomly initialized weight matrix by ${}_0 \ranmat{W}^{(l)}.$ We would like to evaluate 
\begin{equation}
\label{eq:limitiidkernel}
\lim\limits_{m \to \infty}\mathbb{E} \big[ \sigma({}_0 \ranvec{W}^{(l)}_j \cdot \mathbf{\hat{x}}_{(m)})  \sigma({}_0 \ranvec{W}^{(l)}_j \cdot \mathbf{\hat{y}}_{(m)}) \big].
\end{equation}

The following gives a sufficient condition for the central limit theorem (CLT) to apply. Let $x_{(m)i}$ denote the $i$th coordinate of $\mathbf{x}_{(m)}$.

\begin{hyp}
\label{hyp:decay}
$\lim\limits_{m\to \infty} m^{(1/4)} \max_{i=1}^m \frac{ |x_{(m)i}|}{\Vert \mathbf{x}_{(m)} \Vert}$ and $\lim\limits_{m\to \infty} m^{(1/4)} \max_{i=1}^m \frac{ |y_{(m)i}|}{\Vert \mathbf{y}_{(m)} \Vert}$ are both $0$.
\end{hyp}

This condition is easily satisfied since for data points with many non-zero entries, $\Vert \mathbf{x}_{(m)} \Vert$ will grow like $\sqrt{m}$ when compared to $|x_{(m)i}|$. Provided $\mathbb{E}\big[ {}_0 W_{11}^{(l)} \big]=0$ and $\mathbb{E}\big| {}_0 W_{11}^{(l)} \big|^3<\infty$, \cite{tsuchida2017invariance} show that under Hypothesis~\ref{hyp:decay}, $$\sigma \Big( {}_0 \ranvec{W}^{(l)}_1 \cdot \mathbf{\hat{x}}_{(m)} \Big) \sigma \Big( {}_0 \ranvec{W}^{(l)}_1 \cdot \mathbf{\hat{y}}_{(m)} \Big) \xrightarrow{d}\sigma(Z_\mathbf{x})\sigma(Z_\mathbf{y}),$$
$(Z_\mathbf{x}, Z_\mathbf{y}) \sim \mathcal{N}(\ranvec{0}, \Sigma)$ with ${\Sigma = \begin{bmatrix}
1 & \cos\theta^{(0)} \\
\cos\theta^{(0)} & 1 
\end{bmatrix}}.$

Letting $Z_{\mathbf{x}(m)} = {}_0 \ranvec{W}^{(l)}_j \cdot \mathbf{\hat{x}}_{(m)},$ $Z_{\mathbf{y}(m)} = {}_0 \ranvec{W}^{(l)}_j \cdot \mathbf{\hat{y}}_{(m)},$ 
\begin{align*}
\sigma(Z_{\mathbf{x}(m)}) \sigma(Z_{\mathbf{y}(m)}) &\leq |Z_{\mathbf{x}(m)}| |Z_{\mathbf{y}(m)}| \\
&\leq \text{max}\big\{ Z_{\mathbf{x}(m)}^2, Z_{\mathbf{y}(m)}^2 \big\} \\
&\leq Z_{\mathbf{x}(m)}^2 + Z_{\mathbf{y}(m)}^2.
\end{align*}
The integral of the RHS is equal to $2\mathbb{E}\big[{}_0 W_{11}^{(l)} \big] $ and so the limit may be brought inside the integral in~\eqref{eq:limitiidkernel} by Theorem 19 of \cite{royden2010real}\footnote{This is a variation of the usual dominated convergence theorem where the dominating function $g \geq f_m$ is replaced by a dominating \emph{sequence} $g_m \geq f_m$.}. The resulting expectation corresponds to~\eqref{eq:arccosker}.

Figure~\ref{fig:kernelinit} shows the normalized kernels for weight distributions with the PDF
$\prod_{i=1}^{m} \frac{\beta}{2\alpha \Gamma(1/\beta)} e^{-|w_i/\alpha|^\beta}$. This PDF generalizes the isotropic Gaussian PDF ($\beta = 2$) and the Uniform PDF ($\beta \to \infty$).

The CLT result says nothing about the kernel of \emph{trained} networks whose weights are not IID. In \S\ref{sec:kernelimplications} we extend the CLT result to trained networks. We do this by first exploring exchangeability in MLPs.

\section{Exchangeability in MLPs}
\label{sec:exchangeable}
Suppose that for every $l$, the matrix $( {}_0 W_{ji}^{(l)} )_{ji}$ is IID and then the weights evolve according to SGD over $t$ iterations. The index $j$ (which corresponds to the $j$th row of the random weight matrix, or the $j$th neuron in layer $l$) is an arbitrary labeling; one may permute these indices along with the corresponding connection in layer $l+1$ without changing the output of the network or the joint distribution of the weights. 

We show this for $L=3$; the generalization to any $L\geq2$ will be clear. To start our argument, it is obvious that there is full exchangeability of the weights when the network has been randomly initialized with IID weights and has not yet been trained. More restrictively, we have the following.

\begin{obs}
Let $\mathbf{a} \in \mathbb{R}^{n^{(0)}}$ and $\mathbf{b} \in \mathbb{R}^{n^{(3)}}$ be inputs and targets of an MLP, respectively. Suppose that the initial weights in each layer ${}_0 \ranmat{W}^{(l)}$ are IID, and temporarily drop the pre-subscript. Then for any bijective permutations $\pi_1$ and $\pi_2$,
\begin{align*}
 &\Big( \ranvec{a}, \big( W_{ \pi_1(i)h}^{(1)} \big)_{ih}, \big( W_{\pi_2(j)\pi_1(i)}^{(2)} \big)_{ji}, \big( W_{k\pi_2(j)}^{(3)} \big)_{ kj}, \ranvec{b} \Big)  \\
\stackrel{d}{=}&\Big( \ranvec{a}, \big( W_{ih}^{(1)} \big)_{ih}, \big( W_{ji}^{(2)} \big)_{ji}, \big( W_{kj}^{(3)} \big)_{ kj}, \ranvec{b} \Big). \numberthis \label{eq:exchange}
\end{align*}
\end{obs}
This generalizes to any network with one or more hidden layers ($L \geq 2$) because the permutation does not affect the non-exchangeable elements $\mathbf{a}$ and/or $\mathbf{b}$.

Define $g^{(l)}_{qp}$ to be the function that takes $\mathbf{a}$, $\mathbf{b}$ and realizations of $\big( {}_0 \ranmat{W}^{(m)} \big)_{m \in [L]}$ and calculates realizations of ${}_1 W_{qp}^{(l)}$ according to an online (batch size of $1$) backpropagation update rule. Let $\overline{\mathbf{g}}^{(l)}$ be a matrix-valued function, whose $qp$th element is $g^{(l)}_{qp}$.  We have
\begin{alignat*}{2}
\overline{\mathbf{g}}^{(l)} \Big( \mathbf{a}, \mathbf{b}, \big( \ranmat{w}^{(r)} \big)_{r \in [L]} \Big) &= \ranmat{w}^{(l)} - \alpha \frac{\partial E}{\partial \ranmat{w}^{(l)} },
\end{alignat*}
for some cost function $E\Big( \mathbf{a}, \mathbf{b}; \big( \ranmat{w}^{(r)} \big)_{r \in [L]} \Big)$ and step-size $\alpha \in \mathbb{R}$. Denote the LHS of~\eqref{eq:exchange} by $\mathbf{U}$ and the RHS of~\eqref{eq:exchange} by $\mathbf{U}_\pi$. Then by examining the backpropagation equations,
\begin{align*}
\frac{\partial E}{\partial w_{ji}^{(l)}} \Bigg|_{\mathbf{U}_\pi} &= \frac{\partial E}{\partial  w_{\pi_2(j)\pi_1(i)}^{(l)}} \Bigg|_{\mathbf{U}}.
\end{align*}

By the continuous mapping theorem, we may apply $\mathbf{g}^{(2)}$ to both sides of~\eqref{eq:exchange} if $\mathbf{g}^{(2)}$ is almost everywhere (a.e.) continuous. We have (again, temporarily dropping the $0$ pre-subscripts on the weights),
\begin{align*}
 &\overline{\mathbf{g}}^{(2)} \Big( \ranvec{a}, \big( W_{ih}^{(1)} \big)_{ih}, \big( W_{ji}^{(2)} \big)_{ji}, \big( W_{kj}^{(3)} \big)_{ kj }, \ranvec{b} \Big) , \\
\stackrel{d}{=}& \overline{\mathbf{g}}^{(2)} \Big( \ranvec{a}, \big( W_{ \pi_1(i)h}^{(1)} \big)_{ih }, \big( W_{\pi_2(j)\pi_1(i)}^{(2)} \big)_{ji}, \big( W_{k\pi_2(j)}^{(3)} \big)_{ kj  }, \ranvec{b} \Big), \\ 
=& \bigg( g^{(2)}_{\pi_2(q)\pi_1(p)} \Big( \ranvec{a}, \big( W_{ih}^{(1)} \big)_{ih }, \big( W_{ji}^{(2)} \big)_{ji}, \big( W_{kj}^{(3)} \big)_{ kj  }, \ranvec{b} \Big) \bigg)_{qp}, \\
=& \bigg( {}_1 W_{\pi_2(q)\pi_1(p)}^{(2)} \bigg)_{qp}, \numberthis \label{eq:rowex}
\end{align*}
and the first line is equal to $\big( {}_1 W_{qp}^{(2)} \big)_{qp}$. This shows that the rows and columns of ${}_1 \ranmat{W}^{(2)}$ may be permuted without affecting the joint distribution, i.e. ${}_1 \ranmat{W}^{(2)}$ is RCE. ${}_t \ranmat{W}^{(1)}$ is row but not column-exchangeable and ${}_t \ranmat{W}^{(L)}$ is column but not row-exchangeable.

When any batch size is used,~\eqref{eq:rowex} again holds since the inputs $\ranvec{a}$ and $\ranvec{b}$ may be replaced by sets $\{ \ranvec{a}_i \}_{i \leq M}$ and $\{ \ranvec{b}_i \}_{i \leq M}$ where $M$ is the batch-size. One may choose any $M$; if $M$ is the size of the entire finite dataset, this corresponds to gradient descent. We may use any a.e. continuous $\ranmat{g}^{(l)}$ whose evaluation commutes with index permutations in the input (such as SGD, Adam~\citep{kingma2015adam} or RMSprop). Call such an update rule \emph{index commuting}. By redefining $\mathbf{g}^{(2)}$ to calculate the weights at the $t$th iteration of SGD, one can show that ${}_t \ranmat{W}^{(2)}$ is RCE $\forall t$.

\begin{theorem}
\label{thm:rowcolex}
Let $L \geq 3$. Suppose that the initial weights in each layer ${}_0 \ranmat{W}^{(l)}$ are IID. Suppose the network is trained using an index commuting update rule. Then for all $2 \leq l \leq L-1$ and all optimizer iterations $t \geq 0$, the weight matrices ${}_t \ranmat{W}^{(l)}$ are RCE. 

For $L \geq 2$, ${}_t \ranmat{W}^{(1)}$ is row but not column exchangeable and ${}_t \ranmat{W}^{(L)}$ is column but not row exchangeable.
\end{theorem}

\section{Kernels of Trained MLPs}
\label{sec:kernelimplications}
We now extend the results of \S\ref{sec:asymker} to trained networks using the results of \S\ref{sec:exchangeable}.
\subsection{Layer-wise kernel in trained MLPs}
\label{sec:layerkernel}
We examine the limit in $m$ of the layer-wise kernel in layer $l$ for a network with RCE weights. By Theorem~\ref{thm:aldous1}, there exists some measurable function $f$ and some mutually independent $A$ and $\ranvec{B}$ each uniform on $[0,1]$ such that
\begin{align*}
&\phantom{{}={}}\lim_{m \to \infty}\mathbb{E}\Big[ \sigma \Big( {}_t \ranvec{W}^{(l)}_1 \cdot \mathbf{\hat{x}}_{(m)} \Big) \sigma \Big( {}_t \ranvec{W}^{(l)}_1 \cdot \mathbf{\hat{y}}_{(m)} \Big) \Big] \\
&= \lim_{m \to \infty} \int_{[0,1]} k_A(\mathbf{x}_{(m)}, \mathbf{y}_{(m)}) \,d\mu_A, \numberthis \label{eq:intka}
\end{align*}
where $\mu_A$ is the uniform probability measure on $[0,1]$ with $k_A(\mathbf{x}_{(m)}, \mathbf{y}_{(m)})$ given by
\begin{align*}
\int\limits_{[0,1]^{\infty}} \sigma \Big( \mathbf{f}_A(\ranvec{B}) \cdot \mathbf{\hat{x}}_{(m)} \Big) \sigma \Big( \mathbf{f}_A(\ranvec{B}) \cdot \mathbf{\hat{y}}_{(m)} \Big) \,d\mu_{\ranvec{B}},
\end{align*}
where $\big( \ranvec{f}_A(\ranvec{B})_i \big)_i = \big( f(A,B_i) \big)_i$ and $\mu_\ranvec{B}$ is the uniform probability measure on $[0,1]^\infty$. \textbf{For the remainder of the paper we will drop the pre-subscript $t$ denoting the training iteration on the weights.}

Now we bring the limit inside the integral in~\eqref{eq:intka} (again using Theorem 19 of~\cite{royden2010real}). Using the fact that $\sigma(d)\sigma(e) \leq |d||e|\leq \text{max} \{ d^2, e^2\}\leq d^2 + e^2$, $k_A(\mathbf{x}_{(m)}, \mathbf{y}_{(m)})$ is dominated by 
\begin{align*}
&\int\limits_{[0,1]^{\infty}}  \Big( \mathbf{f}_A(\ranvec{B}) \cdot \mathbf{\hat{x}}_{(m)} \Big)^2 \,d\mu_{\ranvec{B}} + \int\limits_{[0,1]^{\infty}}  \Big( \mathbf{f}_A(\ranvec{B}) \cdot \mathbf{\hat{y}}_{(m)}  \Big)^2 \,d\mu_{\ranvec{B}}.
\end{align*}
The first of these terms is given by
\begin{align*}
&\int_{[0,1]}f_A(B_{1})^2\,d\mu_{B_{1}} + \Big(\int_{[0,1]}f_A(B_{1})\,d\mu_{B_{1}}\Big)^2 \sum_{i=1}^m \hat{x}_{(m)i}\sum_{\substack{j=1 \\ j\neq i}}^m \hat{x}_{(m)j}.
\end{align*}
This quantity is integrable with respect to $\mu_A$ for all $m$ if $\mathbb{E} \big[ (W^{(l)}_{11})^2\big]$ and $\mathbb{E} \big[ W^{(l)}_{11} W^{(l)}_{12}\big]$ are finite. Now we may bring the limit inside the integral in~\eqref{eq:intka} provided that $\lim\limits_{m\to\infty}\sum\limits_{i=1}^m \hat{x}_{(m)i}$ and $\lim\limits_{m\to\infty}\sum\limits_{i=1}^m \hat{y}_{(m)i}$ exist and are finite (see Appendix~\ref{app:doublesum}). 

Write $f_A(B_{1})=g_A(B_{1})+E_A$ with $E_A = \mathbb{E}_{B_1}f_A(B_1)$ and $\mathbb{E}_{B_1} g_A(B_{1})=0$. Using $\sigma(a+b) \leq \sigma(a) + \sigma(b),$ $k_A(\mathbf{x}_{(m)}, \mathbf{y}_{(m)})$ is bounded above by
\begin{align*}
&\mathbb{E}_{\ranvec{B}} \Big[\sigma \Big( \mathbf{g}_A(\ranvec{B}) \cdot \mathbf{\hat{x}}_{(m)} \Big) \sigma \Big( \mathbf{g}_A(\ranvec{B}) \cdot \mathbf{\hat{y}}_{(m)} \Big) \Big] +\\
&\mathbb{E}_{\ranvec{B}} \Big[\sigma \big(E_A  L_{\mathbf{y}_{(m)}} \big) \sigma \Big( \mathbf{g}_A(\ranvec{B}) \cdot \mathbf{\hat{x}}_{(m)} \Big) \Big] + \\
&\mathbb{E}_{\ranvec{B}} \Big[\sigma \big(E_A L_{\mathbf{x}_{(m)}} \big) \sigma \Big( \mathbf{g}_A(\ranvec{B}) \cdot \mathbf{\hat{y}}_{(m)} \Big) \Big] + \\
&\mathbb{E}_{\ranvec{B}} \Big[\sigma \big(E_A L_{\mathbf{x}_{(m)}} \big) \sigma \big(E_A L_{\mathbf{y}_{(m)}}\big) \Big],
\end{align*}
where $L_{\mathbf{x}_{(m)}} = \sum_{i=1}^m \hat{x}_{(m)i}$ and $L_{\mathbf{y}_{(m)}} = \sum_{i=1}^m \hat{y}_{(m)i}$. We label each term in the sum in order of appearance as $I_{A(m)1}$, $I_{A(m)2}$, $I_{A(m)3}$ and $I_{A(m)4}$.

As outlined in \S\ref{sec:asymker}, $\lim\limits_{m\to \infty} I_{A(m)1}$ is
$$ \frac{s_A^2}{2\pi} \big( \sin\theta^{(l-1)}+(\pi-\theta^{(l-1)})\cos\theta^{(l-1)}\big),$$
where $s_A^2=\int_{[0,1]}\big( f_A(B_1)-E_A)^2 \,d\mu_{B_1}.$ Integrating with respect to $\mu_A$, 
\begin{align*}
\int_{[0,1]} s_A^2 \,d\mu_A=\mathbb{E}\big[  (W^{(l)}_{11})^2 \big] - \mathbb{E}\Big[  W^{(l)}_{11} W^{(l)}_{12} \Big].
\end{align*}
This implies that $\int\limits_{[0,1]}\lim\limits_{m\to\infty} I_{A(m)1} \,d\mu_A$ is given by 
\begin{align*}
& \frac{1}{2\pi} \Big(\mathbb{E}\big[ (W^{(l)}_{11})^2 \big] - \mathbb{E}\big[ W^{(l)}_{11} W^{(l)}_{12} \big] \Big) \big( \sin\theta^{(l-1)}+(\pi-\theta^{(l-1)})\cos\theta^{(l-1)}\big). \numberthis \label{eq:i1}
\end{align*}

$\mathbb{E}_A I_{(m)2}$ is bounded by (see Appendix~\ref{app:i2bound})
\begin{align}
&\Bigg| L_{\mathbf{y}_{(m)}} \sqrt{\mathbb{E}\big[ W_{11}^{(l)}W_{12}^{(l)}\big]} \sqrt{ \mathbb{E} \big[ (W_{11}^{(l)})^2 \big] - \mathbb{E} \big[W_{11}^{(l)} W_{12}^{(l)} \big]} \Bigg|.
\label{eq:bound1}
\end{align}
A bound for $\mathbb{E}_A I_{(m)3}$ follows from symmetry.

$\mathbb{E}_A  I_{(m)4}$ is bounded by (see Appendix~\ref{app:i4bound})
\begin{equation}
\mathbb{E}\big[ W_{11}^{(l)} W_{12}^{(l)} \big] \big| L_{\mathbf{y}_{(m)}} L_{\mathbf{y}_{(m)}} \big|.
\label{eq:bound2}
\end{equation}

We now observe that the difference between~\eqref{eq:intka} and~\eqref{eq:i1} is bounded by the sum of three terms involving $\mathbb{E}\big[ W_{11}^{(l)} W_{12}^{(l)} \big]$, $\mathbb{E}\big[ ( W_{11}^{(l)})^2 \big]$, $\lim\limits_{m \to \infty} \sum\limits_{i=1}^m \hat{x}_{(m)i}$ and $\lim\limits_{m \to \infty} \sum\limits_{i=1}^m \hat{y}_{(m)i}$. This implies the following.

\begin{prop}
\label{prop:trainedkernel}
Suppose that $L \geq 3$, $2 \leq l \leq L-1$ and $\mathbb{E}\big| W_{11}^{(l)} \big|^3 < \infty,$ $\mathbb{E}\big| W_{11}^{(l)} W_{12}^{(l)} \big| < \infty,$ Hypothesis~\ref{hyp:decay} is satisfied and ${\lim\limits_{m \to \infty} \sum\limits_{i=1}^m \hat{x}_{(m)i}=\lim\limits_{m \to \infty} \sum\limits_{i=1}^m \hat{y}_{(m)i}=0}$. 

Then~\eqref{eq:intka} is given by~\eqref{eq:i1}.
\end{prop}
Note that~\eqref{eq:i1} and~\eqref{eq:arccosker} are the same up to a scaling factor, which cancels out after normalizing.
\subsection{The Ergodic Problem}
Unfortunately,~\eqref{eq:intka} is not necessarily the inner product in feature space of an infinitely wide network. To see this, note that by Theorem~\ref{thm:aldous2},
\begin{align*}
&\phantom{{}={}} \frac{1}{n} \sum_{j=1}^n \sigma \big( \ranvec{W}_j^{(l)} \cdot \mathbf{\hat{x}}_{(m)} \big) \sigma \big( \ranvec{W}_j^{(l)} \cdot \mathbf{\hat{y}}_{(m)}\big) \\
&\stackrel{d}{=} \frac{1}{n} \sum_{j=1}^n \sigma \big(\ranvec{f}_{A\ranvec{C}}(B_j, \ranvec{D}_j) \cdot \mathbf{\hat{x}}_{(m)} \big) \sigma \big(\ranvec{f}_{A\ranvec{C}}(B_j, \ranvec{D}_j) \cdot \mathbf{\hat{y}}_{(m)}\big),
\end{align*}
for some measurable ${\ranvec{f}_{A\ranvec{C}}(B_1,\ranvec{D}_1)=(f(A,B_j,C_i,D_{ji})_i,}$ which converges almost surely to the random variable depending on $A$ and $\ranvec{C}$
\begin{equation}
\label{eq:innerproduct_birkhoff}
\mathbb{E}_{B_1\ranvec{D}_1} \Big[ \sigma \big(\ranvec{f}_{A\ranvec{C}}(B_1, \ranvec{D}_1) \cdot \mathbf{\hat{x}}_{(m)} \big) \sigma \big(\ranvec{f}_{A\ranvec{C}}(B_1, \ranvec{D}_1) \cdot \mathbf{\hat{y}}_{(m)}\big) \Big]
\end{equation}
by the Birkhoff-Khinchin ergodic theorem (see Appendix~\ref{app:birkhoff}). For the purposes of experimenting, we make the following simplifying assumption.

\begin{hyp}
\label{hyp:ergodic}
The following holds:
\begin{align*}
&\phantom{\xrightarrow{p}}\frac{1}{n}\sum_{j=1}^n \sigma \big( \ranvec{W}_j^{(l)} \cdot \hat{\mathbf{x}}_{(m)} \big) \sigma \big(\ranvec{W}_j^{(l)} \cdot \hat{\mathbf{y}}_{(m)} \big)  \xrightarrow{p}\mathbb{E} \Big[ \sigma \big(\ranvec{W}_1^{(l)} \cdot \hat{\mathbf{x}}_{(m)} \big) \sigma \big( \ranvec{W}_1^{(l)} \cdot \hat{\mathbf{y}}_{(m)} \big) \Big].
\end{align*}
\end{hyp}
Hypothesis~\ref{hyp:ergodic} says that taking averages over $j$ of the products of activations ${\sigma \big( \ranvec{W}_j^{(l)} \cdot \hat{\mathbf{x}}_{(m)} \big) \sigma \big(\ranvec{W}_j^{(l)} \cdot \hat{\mathbf{y}}_{(m)} \big)}$ in \emph{one} network is equivalent to taking averages over one fixed neuron of the products of activations in an \emph{ensemble} of independent networks. 

A sufficient condition for Hypothesis~\ref{hyp:ergodic} to hold is that the measure of the product of activations is \emph{ergodic} with respect to the row-shift transformation. This condition is stronger than necessary. In statistical mechanics, an ``approximate ergodicity" applied to \emph{sum functions} is used to compare time averages with phase averages~\citep{khinchin1949mathematical, kurth2014axiomatics}. The Ergodic Problem features heavily in the history of statistical mechanics~\citep{moore2015ergodic}. It is our hope that by  introducing this assumption into the analysis of MLPs, we make further progress towards efforts in connecting neural networks to statistical mechanics~\citep{martin2017rethinking}. In \S\ref{sec:exp} we demonstrate that Hypothesis~\ref{hyp:ergodic} is not inconsistent with our empirical observations.

\section{Experiments}
\label{sec:exp}
We illustrate our results with a subset of all figures. Experiments on other datasets and optimizers are provided in the supplementary material.
\subsection{Experiment 1: Verification of Proposition~~\ref{prop:trainedkernel}}
\label{sec:exp1}
\textbf{Architecture:} We use an MLP autoencoder with $4$ layers and $3072$ neurons in each layer that is trained on CIFAR10~\citep{krizhevsky2009learning} ($32 \times 32 \times 3$ images whose pixel values are normalized to $[0, 1]$) using an $\ell2$ reconstruction error objective. Weights are initialized with a variance of $\frac{2}{n_l}$~\citep{he2015delving}.

\textbf{Method:} In Figure~\ref{fig:exp1} we plot the empirical layer-wise normalized kernel in each layer. The color of the points moves from blue to red as the training iteration $t$ increases. Each sample is generated using Procedure~\ref{pro:sample}. The numerical steps ensure that the desired angle $\theta^{(l-1)}$ is obtained between $\mathbf{x}$ and $\mathbf{y}$. The alphabetical steps ensure that $\sum_{i=1}^m x_i = \sum_{i=1}^m y_i = 0$. 
\begin{algorithm}
\caption{Sample $\theta^{(l-1)}$}
\label{pro:sample}
\hspace*{\algorithmicindent} \textbf{Inputs} Dataset $\mathcal{X}$, \quad $\theta^{(l-1)} \in [0, \pi]$ \\
\hspace*{\algorithmicindent} \textbf{Outputs} $\mathbf{x} \in \mathcal{X}$, \quad $\mathbf{y}$ whose angle to $\mathbf{x}$ is $\theta^{(l-1)}$
\begin{enumerate}
\item Pick an $\mathbf{x}$ uniform-randomly from $\mathcal{X}$. 
\begin{enumerate}
\item Set the last two coordinates of $\mathbf{x}$ to $0$.
\end{enumerate}
\item Generate a random vector $\mathbf{p}$ orthogonal to $\mathbf{x}$: Excluding the last two coordinates of $\mathbf{p}$ which are set to $0$, set all coordinates of $\mathbf{p}$ to zero where $\mathbf{x}$ is non-zero and sample all coordinates of $\mathbf{p}$ from $\mathcal{U}[0,1]$ where $\mathbf{x}$ is zero. Normalize $\mathbf{p}$ to have the same $\ell 2$ norm as $\mathbf{x}$. 
\begin{enumerate}
\item Set the second last coordinate of $\mathbf{x}$ to the negative sum of all coordinates of $\mathbf{x}$.
\item Set the last coordinate of $\mathbf{p}$ to the negative sum of all coordinates of $\mathbf{p}$.
\end{enumerate}
\item Set $\mathbf{y} = \cos\theta^{(l-1)} \mathbf{x} + \sin\theta^{(l-1)} \mathbf{p}$.
\end{enumerate}
\end{algorithm}

Although we do not explore this idea deeply here, step $3$ may be viewed as generating an adversarial perturbation of $\mathbf{x}$ for small $\theta^{(l-1)}$.

\newcommand\scalev{0.15}
\newcommand{\sizespace}{0.15}
\newcommand{\aesize}{0.18}

\renewcommand\scalev{0.15}
\renewcommand{\sizespace}{0.15}
\renewcommand{\aesize}{0.8}

\def\directoryadam{zero_sum/cifar10/adam}
\def\directorysgd{zero_sum/cifar10/sgd}

\def\slurmidsadam{77896, 77899, 77903, 77907}
\def\slurmidssgd{77922}

\def\colorbar{colorbar19000}

\def\figlab{fig:exp1}

\def\figcap{Layer-wise normalized kernels for a trained MLP at iteration $t$, indicated by color. Experimental details outlined in \S\ref{sec:exp1}. Batch-size of $256$ used. \textbf{First 4 columns:} normalized layer-wise kernels in layers $1$ to $4$. \textbf{Fifth column:} normalized kernel for the full network. \textbf{Last column:} sample reconstruction on test data. \textbf{First 5 rows:} network trained using Adam using step size $0.001$, $\beta_1=0.9$, $\beta_2=0.999$, ${\varepsilon=[10^{-16}, 10^{-8}, 10^{-4}, 1]}$. \textbf{Last row:} Network trained using SGD with constant learning rate $0.5$.}

\def\iternum{19000}

\def\filetype{png}
\def\quality{tn}

\ifthenelse{\boolean{figuresbool}}{
\begin{figure*}[t!]
\centering
\foreach \id in \slurmidsadam
{
	\begin{subfigure}[t]{\sizespace \linewidth}
	\includegraphics[scale=\scalev]{figures/\directoryadam/\id/kernel_1_\quality.\filetype} 
	\end{subfigure}
	\begin{subfigure}[t]{\sizespace \linewidth}
	\includegraphics[scale=\scalev]{figures/\directoryadam/\id/kernel_2_\quality.\filetype} 
	\end{subfigure}
	\begin{subfigure}[t]{\sizespace \linewidth}
	\includegraphics[scale=\scalev]{figures/\directoryadam/\id/kernel_3_\quality.\filetype} 
	\end{subfigure}
	\begin{subfigure}[t]{\sizespace \linewidth}
	\includegraphics[scale=\scalev]{figures/\directoryadam/\id/kernel_4_\quality.\filetype} 
	\end{subfigure}
	\begin{subfigure}[t]{\sizespace \linewidth}
	\includegraphics[scale=\scalev]{figures/\directoryadam/\id/kernel_full_\quality.\filetype} 
	\end{subfigure} 
	\begin{subfigure}[t]{\sizespace \linewidth}
	\includegraphics[scale=\aesize, trim=-10 -17 0 0]{figures/\directoryadam/\id/ae\iternum.png}
	\end{subfigure} \newline
}
\foreach \id in \slurmidssgd
{
	\begin{subfigure}[t]{\sizespace \linewidth}
	\includegraphics[scale=\scalev]{figures/\directorysgd/\id/kernel_1_\quality.\filetype} 
	\end{subfigure}
	\begin{subfigure}[t]{\sizespace \linewidth}
	\includegraphics[scale=\scalev]{figures/\directorysgd/\id/kernel_2_\quality.\filetype} 
	\end{subfigure}
	\begin{subfigure}[t]{\sizespace \linewidth}
	\includegraphics[scale=\scalev]{figures/\directorysgd/\id/kernel_3_\quality.\filetype} 
	\end{subfigure}
	\begin{subfigure}[t]{\sizespace \linewidth}
	\includegraphics[scale=\scalev]{figures/\directorysgd/\id/kernel_4_\quality.\filetype} 
	\end{subfigure}
	\begin{subfigure}[t]{\sizespace \linewidth}
	\includegraphics[scale=\scalev]{figures/\directorysgd/\id/kernel_full_\quality.\filetype} 
	\end{subfigure} 
	\begin{subfigure}[t]{\sizespace \linewidth}
	\includegraphics[scale=\aesize, trim=-10 -17 0 0]{figures/\directorysgd/\id/ae\iternum.png}
	\end{subfigure} \newline
}
\begin{subfigure}[b]{0.36\textwidth}
\includegraphics[scale=0.3]{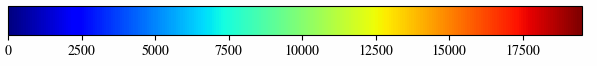}
\end{subfigure}
\caption{\figcap}
\label{\figlab}
\end{figure*}
}
{Figures Here}

\subsection{Experiment 2: Normalized kernels for inputs with non-zero sums}
\label{sec:exp2}
Consider the same architecture as in \S\ref{sec:exp1} with a modified sampling procedure for $\theta^{(l-1)}$: the alphabetical steps of Procedure~\ref{pro:sample} are not performed. This means that the sums $\sum_{i=1}^m \hat{x}_i$ and $\sum_{i=1}^m \hat{y}_i$ are no longer $0$. The bounds~\eqref{eq:bound1} and~\eqref{eq:bound2} are non-zero and so Proposition~\ref{prop:trainedkernel} does not apply. However, \emph{if} $\mathbb{E}\big[  W_{11}^{(l)} W_{12}^{(l)}\big]$ is $0$, these bounds are $0$ and so the result of Proposition~\ref{prop:trainedkernel} holds. Note that
\begin{align*}
\mathbb{E}\big[ W_{11}^{(l)} W_{12}^{(l)}\big] &= \int\limits_{[0,1]^3} f(A,B_1) f(A,B_2) \,d\mu_{B_1B_2A} \\
&= \int\limits_{[0,1]} \Big( \int\limits_{[0,1]} f(A,B_1) \,d\mu_{B_1} \Big)^2  \,d\mu_{A} \\
&=0 \text{ iff } E_A=\int\limits_{[0,1]} f(A,B_1) \,d\mu_{B_1} = 0.
\end{align*}
Also, by the Birkhoff-Khinchin ergodic theorem, 
$$ \frac{1}{n} \sum\limits_{i=1}^n W_{1i}^{(l)} \xrightarrow{a.s.} \int_{[0,1]} f(A,B_1) \,d\mu_{B_1}.$$
(These results are obtained in~\cite{lee1987laws} using different notations). Therefore, for finite $n^{(l)}$ if ${(E^{n^{(l)}})^2 := \max\limits_j \bigg(  \Big( \frac{1}{n^{(l)}} \sum\limits_{i=1}^{n^{(l)}} W_{ji}^{(l)} \Big)^2 \bigg)}$ is ``small'', $\mathbb{E}\big[  W_{11}^{(l)}  W_{12}^{(l)}\big]$ will be ``small''.

We are interested in finding optimizer hyperparameters that result in a change in $(E^{n^{(l)}})^2$ from zero, which in turn results in non-zero bounds~\eqref{eq:bound1} and~\eqref{eq:bound2}. We make the following empirical observations:
\begin{enumerate}
\item When \textbf{Adam} is used, as the hyperparameter $\varepsilon$ decreases there is a sharp change in $(E^{n^{(l)}})^2$ and the mean squared error of the observed normalized kernel to the normalized arc-cosine kernel of degree one measured at iteration $t=19000$. When $(E^{n^{(l)}})^2$ is small the kernel is approximately described by~\eqref{eq:i1}. See Figures~\ref{fig:phase},~\ref{fig:exp2} and Appendices~\ref{app:moreexp},~\ref{app:exp_mnist}. Other optimizers with an $\varepsilon$ hyperparameter, such as \textbf{RMSProp} and \textbf{Nadam}~\citep{dozat2016incorporating} also show this behavior. 
\item For reasonable step sizes $\alpha$ (i.e. those that result in stable training), \textbf{SGD} generally results in smaller $(E^{n^{(l)}})^2$ than Adam, and thus the normalized kernel agrees  more closely with Proposition~\ref{prop:trainedkernel}. See Figures~\ref{fig:phase},~\ref{fig:exp2} and Appendices~\ref{app:moreexp},~\ref{app:exp_mnist}.
\end{enumerate}

\renewcommand\scalev{0.35}
\renewcommand{\sizespace}{0.2}
\begin{figure*}
\centering
\begin{subfigure}[t]{\sizespace \linewidth}
\includegraphics[scale=\scalev]{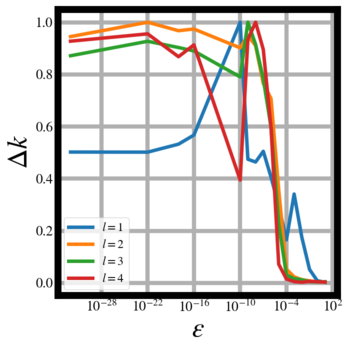} 
\end{subfigure}
\begin{subfigure}[t]{\sizespace \linewidth}
\includegraphics[scale=\scalev]{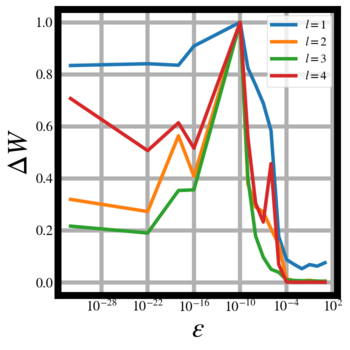} 
\end{subfigure}
\quad
\begin{subfigure}[t]{\sizespace \linewidth}
\includegraphics[scale=\scalev]{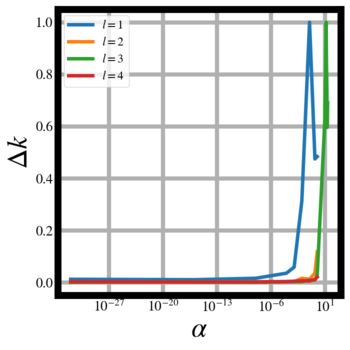} 
\end{subfigure}
\begin{subfigure}[t]{\sizespace \linewidth}
\includegraphics[scale=\scalev]{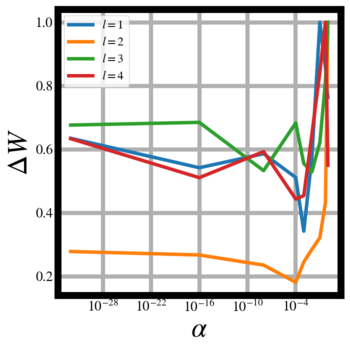} 
\end{subfigure}
\caption{$\Delta k$: Mean squared error of kernel to normalized arc-cosine kernel normalized to between $0$ and $1$. $\Delta W:$ $(E^{n^{(l)}})^2$ normalized to between $0$ and $1$. \textbf{Left:} Network trained with Adam using step size $0.001$, $\beta_1=0.9$, $\beta_2=0.999$ varying $\varepsilon$. \textbf{Right:} Network trained using SGD varying $\alpha$.}
\label{fig:phase}
\end{figure*}

\renewcommand\scalev{0.15}
\renewcommand{\sizespace}{0.15}
\renewcommand{\aesize}{0.8}

\def\directoryadam{nonzero_sum/cifar10/adam}
\def\directorysgd{nonzero_sum/cifar10/sgd}

\def\slurmidsadam{77833, 77836, 77840, 77844}
\def\slurmidssgd{77816}

\def\colorbar{colorbar19000}

\def\figlab{fig:exp2}

\def\figcap{As in Figure~\ref{fig:exp1}, but for inputs with non-zero sums as outlined in \S\ref{sec:exp2}.}

\def\iternum{19000}

\def\filetype{png}
\def\quality{tn}

\ifthenelse{\boolean{figuresbool}}{
\begin{figure*}[t!]
\centering
\foreach \id in \slurmidsadam
{
	\begin{subfigure}[t]{\sizespace \linewidth}
	\includegraphics[scale=\scalev]{figures/\directoryadam/\id/kernel_1_\quality.\filetype} 
	\end{subfigure}
	\begin{subfigure}[t]{\sizespace \linewidth}
	\includegraphics[scale=\scalev]{figures/\directoryadam/\id/kernel_2_\quality.\filetype} 
	\end{subfigure}
	\begin{subfigure}[t]{\sizespace \linewidth}
	\includegraphics[scale=\scalev]{figures/\directoryadam/\id/kernel_3_\quality.\filetype} 
	\end{subfigure}
	\begin{subfigure}[t]{\sizespace \linewidth}
	\includegraphics[scale=\scalev]{figures/\directoryadam/\id/kernel_4_\quality.\filetype} 
	\end{subfigure}
	\begin{subfigure}[t]{\sizespace \linewidth}
	\includegraphics[scale=\scalev]{figures/\directoryadam/\id/kernel_full_\quality.\filetype} 
	\end{subfigure} 
	\begin{subfigure}[t]{\sizespace \linewidth}
	\includegraphics[scale=\aesize, trim=-10 -17 0 0]{figures/\directoryadam/\id/ae\iternum.png}
	\end{subfigure} \newline
}
\foreach \id in \slurmidssgd
{
	\begin{subfigure}[t]{\sizespace \linewidth}
	\includegraphics[scale=\scalev]{figures/\directorysgd/\id/kernel_1_\quality.\filetype} 
	\end{subfigure}
	\begin{subfigure}[t]{\sizespace \linewidth}
	\includegraphics[scale=\scalev]{figures/\directorysgd/\id/kernel_2_\quality.\filetype} 
	\end{subfigure}
	\begin{subfigure}[t]{\sizespace \linewidth}
	\includegraphics[scale=\scalev]{figures/\directorysgd/\id/kernel_3_\quality.\filetype} 
	\end{subfigure}
	\begin{subfigure}[t]{\sizespace \linewidth}
	\includegraphics[scale=\scalev]{figures/\directorysgd/\id/kernel_4_\quality.\filetype} 
	\end{subfigure}
	\begin{subfigure}[t]{\sizespace \linewidth}
	\includegraphics[scale=\scalev]{figures/\directorysgd/\id/kernel_full_\quality.\filetype} 
	\end{subfigure} 
	\begin{subfigure}[t]{\sizespace \linewidth}
	\includegraphics[scale=\aesize, trim=-10 -17 0 0]{figures/\directorysgd/\id/ae\iternum.png}
	\end{subfigure} \newline
}
\begin{subfigure}[b]{0.36\textwidth}
\includegraphics[scale=0.3]{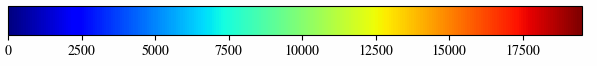}
\end{subfigure}
\caption{\figcap}
\label{\figlab}
\end{figure*}
}
{Figures here}

\section{Discussion and Conclusion}
We identified that the weight matrices in hidden layers of MLPs are RCE. Using this symmetry, we analyzed the kernels of \emph{trained} networks. Specifically, we found that the normalized kernel remains invariant when the inputs have sums over their coordinates of $0$. When the sums are non-zero, a bound which depends on $\mathbb{E}[W^{(l)}_{11}W^{(l)}_{12}]$ applies to the residual of the normalized kernel to the normalized arc-cosine kernel. We derived a measure $(E^{(n^{(l)}})^2$ which, when close to $0$, indicates whether $\mathbb{E}[W^{(l)}_{11}W^{(l)}_{12}]$ is close to $0$ and thus whether the normalized kernel remains approximately invariant during training.

Our theoretical results agree with observations. When empirically comparing optimizers, those which result in small $\mathbb{E}[W^{(l)}_{11}W^{(l)}_{12}]$ have  kernels which follow the normalized arc-cosine kernel and do not change during training. The parameter $\varepsilon$ present in Adam and other optimizers can increase $\mathbb{E}[W^{(l)}_{11}W^{(l)}_{12}]$, leading to qualitatively different kernels to the normalized arc-cosine kernel. Changes in other hyperparameters may change $\mathbb{E}[W^{(l)}_{11}W^{(l)}_{12}]$, although we had difficulty finding instances where changing $\alpha$ in SGD resulted in a kernel that did not roughly match the normalized arc-cosine kernel without also resulting in unstable training.

In contrast with works that analyze the parameter distributions through a small step size approximation of SGD by a stochastic differential equation~\citep{seung1992statistical, watkin1993statistical, martin2017rethinking, chaudhari2018stochastic}, we incorporate very little knowledge of the learning rule into our theory. The result is that our theory is perhaps more general than required. For example, it is interesting to note that our results still hold if we perform (stochastic) gradient \emph{ascent} on the network parameters. We believe our analysis would benefit from including more knowledge of the SGD update rule.

There exists literature concerning the ESD of exchangeable random matrices~\citep{chatterjee2006generalization, adamczak2016circular}. In future work, we would like to apply our results concerning exchangeability to describe the ESD of weight and Hessian matrices in the direction of current work \citep{pennington2017geometry, pennington2017nonlinear}, but without normality assumptions.

\newpage
\newpage
\clearpage
\bibliography{exchangeable}
\bibliographystyle{authordate1}
\newpage

\onecolumn
\renewcommand{\thesubsection}{\Alph{subsection}}
\section*{Supplementary Material}
\subsection{Evaluating the double sum}
\label{app:doublesum}
\begin{align*}
\sum\limits_{i=1}^m \hat{x}_{(m)i} \sum\limits_{j=1}^m \hat{x}_{(m)j} &= \sum\limits_{i=1}^m \hat{x}_{(m)i} \Big( \sum\limits_{\substack{j=1 \\ j\neq i}}^m \hat{x}_{(m)j} + \hat{x}_{(m)i}\Big)\\
&= \sum\limits_{i=1}^m \hat{x}_{(m)i} \sum\limits_{\substack{j=1 \\ j\neq i}}^m \hat{x}_{(m)j} + \hat{x}_{(m)i}^2 \\
&= \sum\limits_{i=1}^m \hat{x}_{(m)i} \sum\limits_{\substack{j=1 \\ j\neq i}}^m \hat{x}_{(m)j} + 1 \\
\Big( \sum\limits_{i=1}^m \hat{x}_{(m)i} \Big)^2 - 1 &=  \sum\limits_{i=1}^m \hat{x}_{(m)i} \sum\limits_{\substack{j=1 \\ j\neq i}}^m \hat{x}_{(m)j}.
\end{align*}

\subsection{A bound for $\mathbb{E}_AI_{A(m)2}$ and $\mathbb{E}_AI_{A(m)3}$}\label{app:i2bound}
\begin{align*}
&\phantom{{}={}}\int\limits_{[0,1]}\int\limits_{[0,1]^\infty} \sigma \Big( E_A \sum_{i=1}^m \hat{y}_{(m)i} \Big) \sigma \Big( \ranvec{g}_A(\ranvec{B}) \cdot \mathbf{\hat{x}} \Big) \, d\mu_{\ranvec{B}} \, d\mu_A \\
&\leq \Bigg| \sum_{i=1}^m \hat{y}_{(m)i} \Bigg| \int\limits_{[0,1]}\int\limits_{[0,1]^\infty} \Big| E_A \Big| \sigma \Big( \ranvec{g}_A(\ranvec{B}) \cdot \mathbf{\hat{x}} \Big) \, d\mu_{\ranvec{B}} \, d\mu_A \\
&= \Bigg| \sum_{i=1}^m \hat{y}_{(m)i} \Bigg| \int\limits_{[0,1]} \Bigg| \int\limits_{[0,1]}f(A,B_1)\,d\mu_{B_1} \Bigg| \int\limits_{[0,1]^\infty}  \sigma \Big( \sum\limits_{i=1}^m g_A(B_i) \hat{x}_i \Big) \, d\mu_{\ranvec{B}} \, d\mu_A \\
& \leq \Bigg| \sum_{i=1}^m \hat{y}_{(m)i} \Bigg| \sqrt{ \int\limits_{[0,1]} \Bigg( \int\limits_{[0,1]}f(A,B_1)\,d\mu_{B_1} \Bigg)^2 \, d\mu_A} \sqrt{ \int\limits_{[0,1]} \Bigg( \int\limits_{[0,1]^\infty}  \sigma \Big( \sum\limits_{i=1}^m g_A(B_i) \hat{x}_i \Big) \, d\mu_{\ranvec{B}} \Bigg)^2 \, d\mu_A } \\
& \leq \Bigg| \sum_{i=1}^m \hat{y}_{(m)i} \Bigg| \sqrt{ \mathbb{E} \big[W_{11}^{(l)} W_{12}^{(l)} \big]} \sqrt{ \int\limits_{[0,1]}  \int\limits_{[0,1]^\infty}  \Big( \sum\limits_{i=1}^m g_A(B_i) \hat{x}_i \Big)^2 \, d\mu_{\ranvec{B}} \, d\mu_A } \\
&= \Bigg| \sum_{i=1}^m \hat{y}_{(m)i} \Bigg| \sqrt{ \mathbb{E} \big[ W_{11}^{(l)} W_{12}^{(l)} \big]} \sqrt{ \int\limits_{[0,1]}  \int\limits_{[0,1]^\infty}  \sum\limits_{i=1}^m g_A(B_i)^2 \hat{x}_i^2   + \sum\limits_{i=1}^m \hat{x}_i g_A(B_i) \sum\limits_{\substack{j=1 \\ j\neq i}}^m \hat{x}_j g_A(B_j) \, d\mu_{\ranvec{B}} \, d\mu_A } \\
&= \Bigg| \sum_{i=1}^m \hat{y}_{(m)i} \Bigg| \sqrt{ \mathbb{E} \big[ W_{11}^{(l)} W_{12}^{(l)} \big]} \sqrt{ \mathbb{E} \big[ ( W_{11}^{(l)})^2 \big] - \mathbb{E} \big[ W_{11}^{(l)} W_{12}^{(l)} \big]}.
\end{align*}
The last line is due to the fact that
\begin{align*}
 \int\limits_{[0,1]^2} g_A(B_i)^2 \,d\mu_A\,d\mu_{B_i} &=  \int\limits_{[0,1]^2} f_A(B_i)^2 - 2f_A(B_i) \int\limits_{[0,1]}f_A(B_k)\,d\mu_{B_k} +  \Big( \int\limits_{[0,1]}f_A(B_k)\,d\mu_{B_k} \Big)^2\,d\mu_{B_i} \,d\mu_A \\
&= \int\limits_{[0,1]} \int\limits_{[0,1]} f_A(B_i)^2 \,d\mu_A \,d\mu_{B_i} - \int\limits_{[0,1]} \Big( \int\limits_{[0,1]}f_A(B_i)\,d\mu_{B_i} \Big)^2\,d\mu_A
\end{align*}
\begin{align*}
\int\limits_{[0,1]^3} g_A(B_i) g_A(B_j) \,d\mu_{B_j} \,d\mu_{B_i} \,d\mu_A &= \int\limits_{[0,1]} \int\limits_{[0,1]} f_A(B_i) \,d\mu_{B_i} \int\limits_{[0,1]} f_A(B_j)  \,d\mu_{B_j} \,d\mu_A - \int\limits_{[0,1]} E_A^2\,d\mu_A = 0
\end{align*}
A bound for $\mathbb{E}_AI_{A(m)3}$ follows from symmetry.

\subsection{A bound for $\mathbb{E}_AI_{A(m)4}$}
\label{app:i4bound}
\begin{align*}
&\phantom{{}={}}\int\limits_{[0,1]} \int\limits_{[0,1]^\infty} \sigma\Big( E_A \sum\limits_{i=1}^m \hat{x}_{(m)i} \Big) \sigma\Big( E_A \sum\limits_{i=1}^m \hat{y}_{(m)i} \Big) \,d\mu_{\ranvec{B}} \,d\mu_A \\
& \leq \int\limits_{[0,1]}  E_A^2 \Big|\sum\limits_{i=1}^m \hat{x}_{(m)i} \Big| \Big|\sum\limits_{i=1}^m \hat{y}_{(m)i} \Big|  \,d\mu_A \\
&= \Big|\sum\limits_{i=1}^m \hat{x}_{(m)i} \Big| \Big|\sum\limits_{i=1}^m \hat{y}_{(m)i} \Big| \int\limits_{[0,1]}  \Bigg( \int\limits_{[0,1]} f(A,B_1) \,d\mu_{B_1} \Bigg)^2 \,d\mu_A \\
&= \Big|\sum\limits_{i=1}^m \hat{x}_{(m)i} \Big| \Big|\sum\limits_{i=1}^m \hat{y}_{(m)i} \Big| \mathbb{E}\Big[  W_{11}^{(l)}  W_{12}^{(l)} \Big].
\end{align*}

\subsection{Convergence of inner product to a conditional expectation}
\label{app:birkhoff}
By Theorem~\ref{thm:aldous2},
\begin{align*}
&\phantom{{}={}} \frac{1}{n} \sum_{j=1}^n \sigma \big( \ranvec{W}_j^{(l)} \cdot \mathbf{\hat{x}}_{(m)} \big) \sigma \big( \ranvec{W}_j^{(l)} \cdot \mathbf{\hat{y}}_{(m)}\big) \\
&\stackrel{d}{=} \frac{1}{n} \sum_{j=1}^n \sigma \big(\ranvec{f}_{A\ranvec{C}}(B_j, \ranvec{D}_j) \cdot \mathbf{\hat{x}}_{(m)} \big) \sigma \big(\ranvec{f}_{A\ranvec{C}}(B_j, \ranvec{D}_j) \cdot \mathbf{\hat{y}}_{(m)}\big).
\end{align*}
for some measurable $\ranvec{f}_{A\ranvec{C}}(B_j,\ranvec{D_j})=\big(f(A,B_j,C_i,D_{ji}) \big)_{i}.$ 

Let $\Omega$ and $\mu$ be the sample space and measure conditional on $A$ and $\ranvec{C}$. Define $T: \Omega \mapsto \Omega$ to be the row-shift transformation so that $T\big( (B_j)_j, (D_{j,i})_{ji} \big)=\big( (B_{j+1})_j, (D_{j+1,i})_{ji} \big)$ and note that $T$ is measure-preserving and ergodic with respect to $\mu$. Let $T^k$ denote $T$ composed $k$ times, with $T^0$ being the identity.

Let $\omega = \big( (B_j)_j, (D_{j,i})_{ji} \big)$. Define $F$, which returns the first row of $\omega$. That is, $F\big( (B_j)_j, (D_{j,i})_{ji} \big) = \big( B_1, (D_{1i})_i \big)$. Then
\begin{align*}
&\phantom{{}={}}\frac{1}{n} \sum_{j=1}^n \sigma \big(\ranvec{f}_{A\ranvec{C}}(B_j, \ranvec{D}_j) \cdot \mathbf{\hat{x}}_{(m)} \big) \sigma \big(\ranvec{f}_{A\ranvec{C}}(B_j, \ranvec{D}_j) \cdot \mathbf{\hat{y}}_{(m)}\big) \\
&=\frac{1}{n} \sum_{j=0}^{n-1} \sigma \Big(\ranvec{f}_{A\ranvec{C}}\big( F(T^j \omega) \big) \cdot \mathbf{\hat{x}}_{(m)} \Big) \sigma \Big(\ranvec{f}_{A\ranvec{C}}\big( F(T^j \omega) \big) \cdot \mathbf{\hat{y}}_{(m)}\Big)
\end{align*}
which converges almost surely to
\begin{equation*}
\int_\Omega \sigma \big(\ranvec{f}_{A\ranvec{C}}(B_1, \ranvec{D}_1) \cdot \mathbf{\hat{x}}_{(m)} \big) \sigma \big(\ranvec{f}_{A\ranvec{C}}(B_1, \ranvec{D}_1) \cdot \mathbf{\hat{y}}_{(m)}\big) \,d\mu
\end{equation*}
by the Birkhoff-Khinchin ergodic theorem. 
\newpage

\subsection{More experiments on CIFAR-10}
\label{app:moreexp}
\subsubsection{SGD with inputs having non-zero sums}
In Figure~\ref{fig:sgdkernels} we show the kernels, samples and learning curves for the networks trained with SGD with a batch size of $256$ and different learning rates.

\renewcommand\scalev{0.13}
\renewcommand{\sizespace}{0.13}
\renewcommand{\aesize}{0.75}

\def\directory{nonzero_sum/cifar10/sgd}

\def\slurmids{77811, 77812, 77813, 77814, 77815, 77816, 77817, 77818, 77819}

\def\colorbar{colorbar19000}

\def\figlab{fig:sgdkernels}

\def\figcap{\textbf{SGD.} Inputs with non-zero sums as outlined in \S\ref{sec:exp2}. \textbf{Top to bottom:} Learning rate ${\alpha = [10^{-8}, 10^{-4}, 10^{-3}, 10^{-2}, 0.1, 0.5,  1, 10, 14]}$.}

\def\iternum{19000}

\def\filetype{png}
\def\quality{tn}

\ifthenelse{\boolean{figuresbool}}{
\begin{figure}[!htb]
\centering
\foreach \id in \slurmids
{
	\begin{subfigure}[t]{\sizespace \linewidth}
	\includegraphics[scale=\scalev]{figures/\directory/\id/kernel_1_\quality.\filetype} 
	\end{subfigure}
	\begin{subfigure}[t]{\sizespace \linewidth}
	\includegraphics[scale=\scalev]{figures/\directory/\id/kernel_2_\quality.\filetype} 
	\end{subfigure}
	\begin{subfigure}[t]{\sizespace \linewidth}
	\includegraphics[scale=\scalev]{figures/\directory/\id/kernel_3_\quality.\filetype} 
	\end{subfigure}
	\begin{subfigure}[t]{\sizespace \linewidth}
	\includegraphics[scale=\scalev]{figures/\directory/\id/kernel_4_\quality.\filetype} 
	\end{subfigure}
	\begin{subfigure}[t]{\sizespace \linewidth}
	\includegraphics[scale=\scalev]{figures/\directory/\id/kernel_full_\quality.\filetype} 
	\end{subfigure} 
	\begin{subfigure}[t]{\sizespace \linewidth}
	\includegraphics[scale=\scalev]{figures/\directory/\id/metric_loss_\quality.\filetype} 
	\end{subfigure}
	\begin{subfigure}[t]{\sizespace \linewidth}
	\includegraphics[scale=\aesize, trim=-10 -14 0 0]{figures/\directory/\id/ae\iternum.png}
	\end{subfigure} \newline
}
\begin{subfigure}[b]{0.36\textwidth}
\includegraphics[scale=0.3]{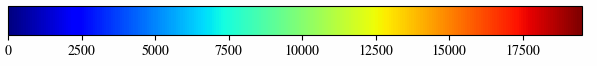}
\end{subfigure}
\caption{\figcap}
\label{\figlab}
\end{figure}
}
{Figures here}

\subsubsection{Adam with inputs having non-zero sums}
In Figure~\ref{fig:adamkernels_a} we show the kernels, samples and learning curves for the networks trained with Adam with a batch size of $256$ and different values of $\varepsilon$. Other hyperparameter values are fixed at $\alpha=0.002$, $\beta_1=0.9$ and $\beta_=0.999$.

\renewcommand\scalev{0.13}
\renewcommand{\sizespace}{0.13}
\renewcommand{\aesize}{0.75}

\def\directory{nonzero_sum/cifar10/adam}

\def\slurmids{77830, 77831, 77832, 77833, 77834, 77835, 77836, 77837}

\def\colorbar{colorbar19000}

\def\figlab{fig:adamkernels_a}

\def\figcap{\textbf{Adam:} Inputs with non-zero sums as outlined in \S\ref{sec:exp2}. \textbf{Top to bottom:} ${\varepsilon = [10^{-32}, 10^{-22}, 10^{-18}, 10^{-16}, 10^{-10}, 10^{-9}, 10^{-8}, 10^{-7}]}$. Figure continues over page...}

\def\iternum{19000}

\def\filetype{png}
\def\quality{tn}

\ifthenelse{\boolean{figuresbool}}{
\begin{figure}[!htb]
\centering
\foreach \id in \slurmids
{
	\begin{subfigure}[t]{\sizespace \linewidth}
	\includegraphics[scale=\scalev]{figures/\directory/\id/kernel_1_\quality.\filetype} 
	\end{subfigure}
	\begin{subfigure}[t]{\sizespace \linewidth}
	\includegraphics[scale=\scalev]{figures/\directory/\id/kernel_2_\quality.\filetype} 
	\end{subfigure}
	\begin{subfigure}[t]{\sizespace \linewidth}
	\includegraphics[scale=\scalev]{figures/\directory/\id/kernel_3_\quality.\filetype} 
	\end{subfigure}
	\begin{subfigure}[t]{\sizespace \linewidth}
	\includegraphics[scale=\scalev]{figures/\directory/\id/kernel_4_\quality.\filetype} 
	\end{subfigure}
	\begin{subfigure}[t]{\sizespace \linewidth}
	\includegraphics[scale=\scalev]{figures/\directory/\id/kernel_full_\quality.\filetype} 
	\end{subfigure} 
	\begin{subfigure}[t]{\sizespace \linewidth}
	\includegraphics[scale=\scalev]{figures/\directory/\id/metric_loss_\quality.\filetype} 
	\end{subfigure}
	\begin{subfigure}[t]{\sizespace \linewidth}
	\includegraphics[scale=\aesize, trim=-10 -14 0 0]{figures/\directory/\id/ae\iternum.png}
	\end{subfigure} \newline
}
\begin{subfigure}[b]{0.36\textwidth}
\includegraphics[scale=0.3]{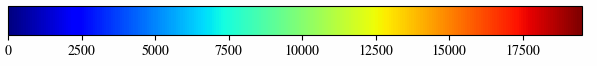}
\end{subfigure}
\caption{\figcap}
\label{\figlab}
\end{figure}
}
{Figures here}

\renewcommand{\thefigure}{\arabic{figure} (Cont.)}
\addtocounter{figure}{-1}

\renewcommand\scalev{0.13}
\renewcommand{\sizespace}{0.13}
\renewcommand{\aesize}{0.75}

\def\directory{nonzero_sum/cifar10/adam}

\def\slurmids{77838, 77839, 77840, 77841, 77842, 77843, 77844}

\def\colorbar{colorbar19000}

\def\figlab{fig:adamkernels_b}

\def\figcap{\textbf{Adam:} Inputs with non-zero sums as outlined in \S\ref{sec:exp2}. \textbf{Top to bottom:} ${\varepsilon = [10^{-6}, 10^{-5}, 10^{-4}, 10^{-3}, 10^{-2}, 10^{-1}, 1]}$.}

\def\iternum{19000}

\def\filetype{png}
\def\quality{tn}

\ifthenelse{\boolean{figuresbool}}{
\begin{figure}[!htb]
\centering
\foreach \id in \slurmids
{
	\begin{subfigure}[t]{\sizespace \linewidth}
	\includegraphics[scale=\scalev]{figures/\directory/\id/kernel_1_\quality.\filetype} 
	\end{subfigure}
	\begin{subfigure}[t]{\sizespace \linewidth}
	\includegraphics[scale=\scalev]{figures/\directory/\id/kernel_2_\quality.\filetype} 
	\end{subfigure}
	\begin{subfigure}[t]{\sizespace \linewidth}
	\includegraphics[scale=\scalev]{figures/\directory/\id/kernel_3_\quality.\filetype} 
	\end{subfigure}
	\begin{subfigure}[t]{\sizespace \linewidth}
	\includegraphics[scale=\scalev]{figures/\directory/\id/kernel_4_\quality.\filetype} 
	\end{subfigure}
	\begin{subfigure}[t]{\sizespace \linewidth}
	\includegraphics[scale=\scalev]{figures/\directory/\id/kernel_full_\quality.\filetype} 
	\end{subfigure} 
	\begin{subfigure}[t]{\sizespace \linewidth}
	\includegraphics[scale=\scalev]{figures/\directory/\id/metric_loss_\quality.\filetype} 
	\end{subfigure}
	\begin{subfigure}[t]{\sizespace \linewidth}
	\includegraphics[scale=\aesize, trim=-10 -14 0 0]{figures/\directory/\id/ae\iternum.png}
	\end{subfigure} \newline
}
\begin{subfigure}[b]{0.36\textwidth}
\includegraphics[scale=0.3]{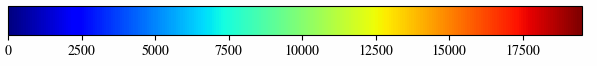}
\end{subfigure}
\caption{\figcap}
\label{\figlab}
\end{figure}
}
{Figures here}

\renewcommand{\thefigure}{\arabic{figure}}

\subsubsection{RMSProp with inputs having non-zero sums}
We search through different values of $\varepsilon$. Figure~\ref{fig:phasermsprop} shows $\Delta W$ and $\Delta k$ as $\varepsilon$ varies. Other hyperparameter values are fixed at $\alpha=0.001$, $\rho=0.9$, no decay and batch size of $256$.

\renewcommand\scalev{0.4}
\renewcommand{\sizespace}{0.45}
\begin{figure}[H]
\centering
\includegraphics[scale=\scalev]{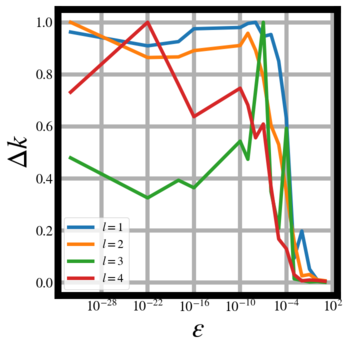} 
\includegraphics[scale=\scalev]{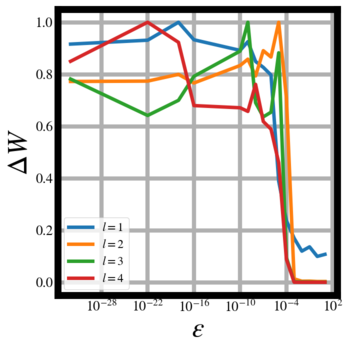} 
\caption{\textbf{RMSProp.} \textbf{Left:} Mean squared error of kernel to normalized arc-cosine kernel and \textbf{Right:} $(E^{n^{(l)}})^2$ as measured against $\varepsilon$. }
\label{fig:phasermsprop}
\end{figure}

In Figure~\ref{fig:rmspropkernels_a} we show the kernels, samples and learning curves for the networks trained using RMSProp.

\renewcommand\scalev{0.13}
\renewcommand{\sizespace}{0.13}
\renewcommand{\aesize}{0.75}

\def\directory{nonzero_sum/cifar10/rmsprop}

\def\slurmids{77866, 77867, 77868, 77869, 77870, 77871, 77872, 77873}

\def\colorbar{colorbar19000}

\def\figlab{fig:rmspropkernels_a}

\def\figcap{\textbf{RMSProp.} Inputs with non-zero sums as outlined in \S\ref{sec:exp2}. \textbf{Top to bottom:} ${\varepsilon = [10^{-32}, 10^{-22}, 10^{-18}, 10^{-16}, 10^{-10}, 10^{-9}, 10^{-8}, 10^{-7}]}$. Figure continues over page...}

\def\iternum{19000}

\def\filetype{png}
\def\quality{tn}

\ifthenelse{\boolean{figuresbool}}{
\begin{figure}[!htb]
\centering
\foreach \id in \slurmids
{
	\begin{subfigure}[t]{\sizespace \linewidth}
	\includegraphics[scale=\scalev]{figures/\directory/\id/kernel_1_\quality.\filetype} 
	\end{subfigure}
	\begin{subfigure}[t]{\sizespace \linewidth}
	\includegraphics[scale=\scalev]{figures/\directory/\id/kernel_2_\quality.\filetype} 
	\end{subfigure}
	\begin{subfigure}[t]{\sizespace \linewidth}
	\includegraphics[scale=\scalev]{figures/\directory/\id/kernel_3_\quality.\filetype} 
	\end{subfigure}
	\begin{subfigure}[t]{\sizespace \linewidth}
	\includegraphics[scale=\scalev]{figures/\directory/\id/kernel_4_\quality.\filetype} 
	\end{subfigure}
	\begin{subfigure}[t]{\sizespace \linewidth}
	\includegraphics[scale=\scalev]{figures/\directory/\id/kernel_full_\quality.\filetype} 
	\end{subfigure} 
	\begin{subfigure}[t]{\sizespace \linewidth}
	\includegraphics[scale=\scalev]{figures/\directory/\id/metric_loss_\quality.\filetype} 
	\end{subfigure}
	\begin{subfigure}[t]{\sizespace \linewidth}
	\includegraphics[scale=\aesize, trim=-10 -14 0 0]{figures/\directory/\id/ae\iternum.png}
	\end{subfigure} \newline
}
\begin{subfigure}[b]{0.36\textwidth}
\includegraphics[scale=0.3]{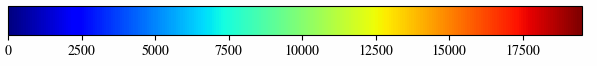}
\end{subfigure}
\caption{\figcap}
\label{\figlab}
\end{figure}
}
{Figures here}

\renewcommand{\thefigure}{\arabic{figure} (Cont.)}
\addtocounter{figure}{-1}

\renewcommand\scalev{0.13}
\renewcommand{\sizespace}{0.13}
\renewcommand{\aesize}{0.75}

\def\directory{nonzero_sum/cifar10/rmsprop}

\def\slurmids{77874, 77875, 77876, 77877, 77878, 77879, 77880}

\def\colorbar{colorbar19000}

\def\figlab{fig:rmspropkernels_b}

\def\figcap{\textbf{RMSProp.} Inputs with non-zero sums as outlined in \S\ref{sec:exp2}. \textbf{Top to bottom:} ${\varepsilon = [10^{-6}, 10^{-5}, 10^{-4}, 10^{-3}, 10^{-2}, 10^{-1}, 1]}$.}

\def\iternum{19000}

\def\filetype{png}
\def\quality{tn}

\ifthenelse{\boolean{figuresbool}}{
\begin{figure}[!htb]
\centering
\foreach \id in \slurmids
{
	\begin{subfigure}[t]{\sizespace \linewidth}
	\includegraphics[scale=\scalev]{figures/\directory/\id/kernel_1_\quality.\filetype} 
	\end{subfigure}
	\begin{subfigure}[t]{\sizespace \linewidth}
	\includegraphics[scale=\scalev]{figures/\directory/\id/kernel_2_\quality.\filetype} 
	\end{subfigure}
	\begin{subfigure}[t]{\sizespace \linewidth}
	\includegraphics[scale=\scalev]{figures/\directory/\id/kernel_3_\quality.\filetype} 
	\end{subfigure}
	\begin{subfigure}[t]{\sizespace \linewidth}
	\includegraphics[scale=\scalev]{figures/\directory/\id/kernel_4_\quality.\filetype} 
	\end{subfigure}
	\begin{subfigure}[t]{\sizespace \linewidth}
	\includegraphics[scale=\scalev]{figures/\directory/\id/kernel_full_\quality.\filetype} 
	\end{subfigure} 
	\begin{subfigure}[t]{\sizespace \linewidth}
	\includegraphics[scale=\scalev]{figures/\directory/\id/metric_loss_\quality.\filetype} 
	\end{subfigure}
	\begin{subfigure}[t]{\sizespace \linewidth}
	\includegraphics[scale=\aesize, trim=-10 -14 0 0]{figures/\directory/\id/ae\iternum.png}
	\end{subfigure} \newline
}
\begin{subfigure}[b]{0.36\textwidth}
\includegraphics[scale=0.3]{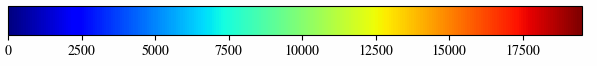}
\end{subfigure}
\caption{\figcap}
\label{\figlab}
\end{figure}
}
{Figures here}

\renewcommand{\thefigure}{\arabic{figure}}

\subsubsection{Nadam with inputs having non-zero sums}
We search through different values of $\varepsilon$. Figure~\ref{fig:phasenadam} shows $\Delta W$ and $\Delta k$ as $\varepsilon$ varies. Other hyperparameter values are fixed at $\alpha=0.002$, $\beta_1=0.9$, $\beta_=0.999$, a decay constant of $0.004$ and batch size of $256$. Note the small values of $\Delta W$ correspond with small values of $\Delta k$.
\renewcommand\scalev{0.4}
\renewcommand{\sizespace}{0.45}
\begin{figure}[H]
\centering
\includegraphics[scale=\scalev]{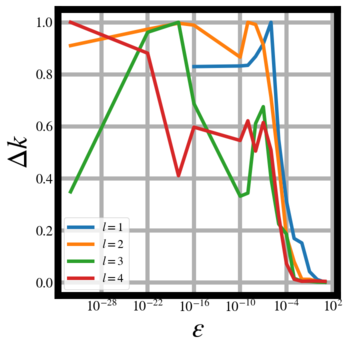} 
\includegraphics[scale=\scalev]{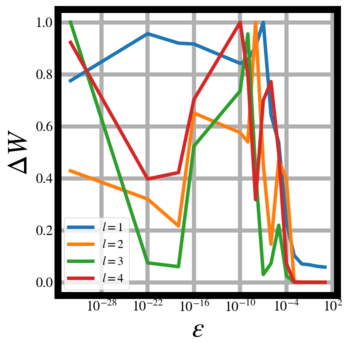} 
\caption{\textbf{Nadam.} \textbf{Left:} Mean squared error of kernel to normalized arc-cosine kernel and \textbf{Right:} $\Delta W$ as measured against $\varepsilon$. }
\label{fig:phasenadam}
\end{figure}

In Figure~\ref{fig:nadamkernels_a} we show the kernels, samples and learning curves for the networks trained using Nadam.

\renewcommand\scalev{0.13}
\renewcommand{\sizespace}{0.13}
\renewcommand{\aesize}{0.75}

\def\directory{nonzero_sum/cifar10/nadam}

\def\slurmids{77847, 77848, 77849, 77850, 77851, 77852, 77853, 77854}

\def\colorbar{colorbar19000}

\def\figlab{fig:nadamkernels_a}

\def\figcap{\textbf{Nadam:} Inputs with non-zero sums as outlined in \S\ref{sec:exp2}. \textbf{Top to bottom:} ${\varepsilon = [10^{-32}, 10^{-22}, 10^{-18}, 10^{-16}, 10^{-10}, 10^{-9}, 10^{-8}, 10^{-7}]}$. Figure continues over page...}

\def\iternum{19000}

\def\filetype{png}
\def\quality{tn}

\ifthenelse{\boolean{figuresbool}}{
\begin{figure}[!htb]
\centering
\foreach \id in \slurmids
{
	\begin{subfigure}[t]{\sizespace \linewidth}
	\includegraphics[scale=\scalev]{figures/\directory/\id/kernel_1_\quality.\filetype} 
	\end{subfigure}
	\begin{subfigure}[t]{\sizespace \linewidth}
	\includegraphics[scale=\scalev]{figures/\directory/\id/kernel_2_\quality.\filetype} 
	\end{subfigure}
	\begin{subfigure}[t]{\sizespace \linewidth}
	\includegraphics[scale=\scalev]{figures/\directory/\id/kernel_3_\quality.\filetype} 
	\end{subfigure}
	\begin{subfigure}[t]{\sizespace \linewidth}
	\includegraphics[scale=\scalev]{figures/\directory/\id/kernel_4_\quality.\filetype} 
	\end{subfigure}
	\begin{subfigure}[t]{\sizespace \linewidth}
	\includegraphics[scale=\scalev]{figures/\directory/\id/kernel_full_\quality.\filetype} 
	\end{subfigure} 
	\begin{subfigure}[t]{\sizespace \linewidth}
	\includegraphics[scale=\scalev]{figures/\directory/\id/metric_loss_\quality.\filetype} 
	\end{subfigure}
	\begin{subfigure}[t]{\sizespace \linewidth}
	\includegraphics[scale=\aesize, trim=-10 -14 0 0]{figures/\directory/\id/ae\iternum.png}
	\end{subfigure} \newline
}
\begin{subfigure}[b]{0.36\textwidth}
\includegraphics[scale=0.3]{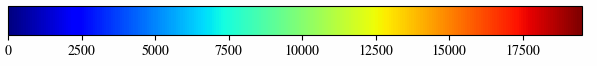}
\end{subfigure}
\caption{\figcap}
\label{\figlab}
\end{figure}
}
{Figures here}

\renewcommand{\thefigure}{\arabic{figure} (Cont.)}
\addtocounter{figure}{-1}

\renewcommand\scalev{0.13}
\renewcommand{\sizespace}{0.13}
\renewcommand{\aesize}{0.75}

\def\directory{nonzero_sum/cifar10/nadam}

\def\slurmids{77855, 77856, 77857, 77858, 77859, 77860, 77861}

\def\colorbar{colorbar19000}

\def\figlab{fig:nadamkernels_b}

\def\figcap{\textbf{Nadam:} Inputs with non-zero sums as outlined in \S\ref{sec:exp2}. \textbf{Top to bottom:} ${\varepsilon = [10^{-6}, 10^{-5}, 10^{-4}, 10^{-3}, 10^{-2}, 10^{-1}, 1]}$.}

\def\iternum{19000}

\def\filetype{png}
\def\quality{tn}

\ifthenelse{\boolean{figuresbool}}{
\begin{figure}[!htb]
\centering
\foreach \id in \slurmids
{
	\begin{subfigure}[t]{\sizespace \linewidth}
	\includegraphics[scale=\scalev]{figures/\directory/\id/kernel_1_\quality.\filetype} 
	\end{subfigure}
	\begin{subfigure}[t]{\sizespace \linewidth}
	\includegraphics[scale=\scalev]{figures/\directory/\id/kernel_2_\quality.\filetype} 
	\end{subfigure}
	\begin{subfigure}[t]{\sizespace \linewidth}
	\includegraphics[scale=\scalev]{figures/\directory/\id/kernel_3_\quality.\filetype} 
	\end{subfigure}
	\begin{subfigure}[t]{\sizespace \linewidth}
	\includegraphics[scale=\scalev]{figures/\directory/\id/kernel_4_\quality.\filetype} 
	\end{subfigure}
	\begin{subfigure}[t]{\sizespace \linewidth}
	\includegraphics[scale=\scalev]{figures/\directory/\id/kernel_full_\quality.\filetype} 
	\end{subfigure} 
	\begin{subfigure}[t]{\sizespace \linewidth}
	\includegraphics[scale=\scalev]{figures/\directory/\id/metric_loss_\quality.\filetype} 
	\end{subfigure}
	\begin{subfigure}[t]{\sizespace \linewidth}
	\includegraphics[scale=\aesize, trim=-10 -14 0 0]{figures/\directory/\id/ae\iternum.png}
	\end{subfigure} \newline
}
\begin{subfigure}[b]{0.36\textwidth}
\includegraphics[scale=0.3]{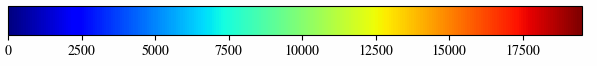}
\end{subfigure}
\caption{\figcap}
\label{\figlab}
\end{figure}
}
{Figures here}

\renewcommand{\thefigure}{\arabic{figure}}

\FloatBarrier
\subsection{More experiments on MNIST}
\label{app:exp_mnist}
\subsection{Adam with inputs having zero sums}
In Figure~\ref{fig:adamkernels_zero} we show the kernels, samples and learning curves for the networks trained with Adam with a batch size of $256$ and different values of $\varepsilon$. Other hyperparameter values are fixed at $\alpha=0.002$, $\beta_1=0.9$ and $\beta_=0.999$.

\renewcommand\scalev{0.13}
\renewcommand{\sizespace}{0.13}
\renewcommand{\aesize}{0.8}

\def\directoryadam{zero_sum/mnist/adam}

\def\slurmids{77750, 77753, 77756, 77760, 77762, 77763, 77764, 77765}

\def\colorbar{colorbar23000}

\def\figlab{fig:adamkernels_zero}

\def\figcap{\textbf{Adam:} Inputs with zero sums as outlined in \S\ref{sec:exp1}, but with MNIST data. \textbf{Top to bottom:} $\varepsilon=[10^{-32}, 10^{-16}, 10^{-8}, 10^{-4}, 10^{-2}, 10^{-1}, 1, 10]$.}

\def\iternum{23000}

\def\filetype{png}
\def\quality{tn}

\ifthenelse{\boolean{figuresbool}}{
\begin{figure*}[t!]
\centering
\foreach \id in \slurmids
{
	\begin{subfigure}[t]{\sizespace \linewidth}
	\includegraphics[scale=\scalev]{figures/\directoryadam/\id/kernel_1_\quality.\filetype} 
	\end{subfigure}
	\begin{subfigure}[t]{\sizespace \linewidth}
	\includegraphics[scale=\scalev]{figures/\directoryadam/\id/kernel_2_\quality.\filetype} 
	\end{subfigure}
	\begin{subfigure}[t]{\sizespace \linewidth}
	\includegraphics[scale=\scalev]{figures/\directoryadam/\id/kernel_3_\quality.\filetype} 
	\end{subfigure}
	\begin{subfigure}[t]{\sizespace \linewidth}
	\includegraphics[scale=\scalev]{figures/\directoryadam/\id/kernel_4_\quality.\filetype} 
	\end{subfigure}
	\begin{subfigure}[t]{\sizespace \linewidth}
	\includegraphics[scale=\scalev]{figures/\directoryadam/\id/kernel_full_\quality.\filetype} 
	\end{subfigure} 
	\begin{subfigure}[t]{\sizespace \linewidth}
	\includegraphics[scale=\scalev]{figures/\directoryadam/\id/metric_loss_\quality.\filetype} 
	\end{subfigure}
	\begin{subfigure}[t]{\sizespace \linewidth}
	\includegraphics[scale=\aesize, trim=-10 -14 0 0]{figures/\directoryadam/\id/ae\iternum.png}
	\end{subfigure} \newline
}
\begin{subfigure}[b]{0.36\textwidth}
\includegraphics[scale=0.3]{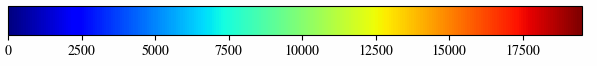}
\end{subfigure}
\caption{\figcap}
\label{\figlab}
\end{figure*}
}
{Figures Here}

\subsubsection{SGD with inputs having non-zero sums}
Figure~\ref{fig:phasesgdmnist} shows $\Delta W$ and $\Delta k$ as $\alpha$ varies. A constant batch size of $256$ was used. 
 
\renewcommand\scalev{0.4}
\renewcommand{\sizespace}{0.45}
\begin{figure}[H]
\centering
\includegraphics[scale=\scalev]{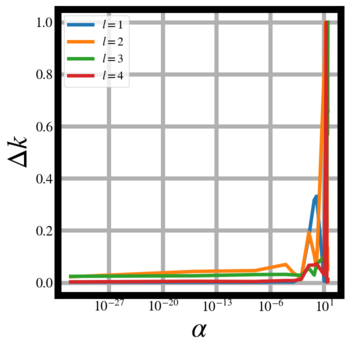} 
\includegraphics[scale=\scalev]{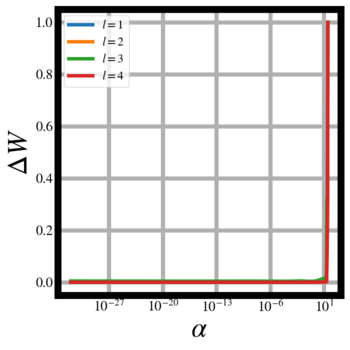} 
\caption{\textbf{SGD.} \textbf{Left:} Mean squared error of kernel to normalized arc-cosine kernel and \textbf{Right:} $(E^{n^{(l)}})^2$ as measured against $\alpha$. }
\label{fig:phasesgdmnist}
\end{figure}

In Figure~\ref{fig:sgdkernelsmnist} we show the kernels, samples and learning curves for the networks trained with SGD with a batch size of $256$ and different learning rates.

\renewcommand\scalev{0.13}
\renewcommand{\sizespace}{0.13}
\renewcommand{\aesize}{0.8}

\def\directory{nonzero_sum/mnist/sgd}

\def\slurmids{77773, 77774, 77775, 77777, 77778, 77781, 77782, 77783, 77784}

\def\colorbar{colorbar23000}

\def\figlab{fig:sgdkernelsmnist}

\def\figcap{\textbf{SGD.} Inputs with non-zero sums as outlined in \S\ref{sec:exp2}, but with MNIST data. \textbf{Top to bottom:} Learning rate ${\alpha = [10^{-3}, 10^{-2}, 10^{-1}, 1, 10, 18, 20, 22, 24]}$.}

\def\iternum{23000}

\def\filetype{png}
\def\quality{tn}

\ifthenelse{\boolean{figuresbool}}{
\begin{figure}[!htb]
\centering
\foreach \id in \slurmids
{
	\begin{subfigure}[t]{\sizespace \linewidth}
	\includegraphics[scale=\scalev]{figures/\directory/\id/kernel_1_\quality.\filetype} 
	\end{subfigure}
	\begin{subfigure}[t]{\sizespace \linewidth}
	\includegraphics[scale=\scalev]{figures/\directory/\id/kernel_2_\quality.\filetype} 
	\end{subfigure}
	\begin{subfigure}[t]{\sizespace \linewidth}
	\includegraphics[scale=\scalev]{figures/\directory/\id/kernel_3_\quality.\filetype} 
	\end{subfigure}
	\begin{subfigure}[t]{\sizespace \linewidth}
	\includegraphics[scale=\scalev]{figures/\directory/\id/kernel_4_\quality.\filetype} 
	\end{subfigure}
	\begin{subfigure}[t]{\sizespace \linewidth}
	\includegraphics[scale=\scalev]{figures/\directory/\id/kernel_full_\quality.\filetype} 
	\end{subfigure} 
	\begin{subfigure}[t]{\sizespace \linewidth}
	\includegraphics[scale=\scalev]{figures/\directory/\id/metric_loss_\quality.\filetype} 
	\end{subfigure}
	\begin{subfigure}[t]{\sizespace \linewidth}
	\includegraphics[scale=\aesize, trim=-10 -14 0 0]{figures/\directory/\id/ae\iternum.png}
	\end{subfigure} \newline
}
\begin{subfigure}[b]{0.36\textwidth}
\includegraphics[scale=0.3]{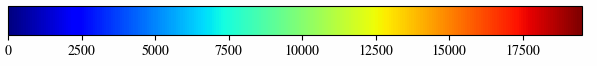}
\end{subfigure}
\caption{\figcap}
\label{\figlab}
\end{figure}
}
{Figures here}

\subsubsection{Adam with inputs having non-zero sums}
We search through different values of $\varepsilon$. Figure~\ref{fig:phaseadammnist} shows $\Delta W$ and $\Delta k$ as $\varepsilon$ varies. Other hyperparameter values are fixed at $\alpha = 0.001$, $\beta_1 = 0.9$, $\beta_2=0.999$, no decay and a batch size of $256$.

\renewcommand\scalev{0.4}
\renewcommand{\sizespace}{0.45}
\begin{figure}[H]
\centering
\includegraphics[scale=\scalev]{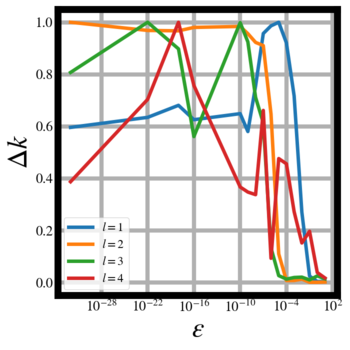} 
\includegraphics[scale=\scalev]{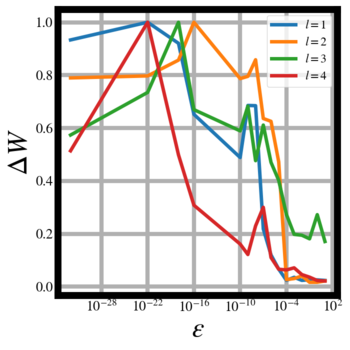} 
\caption{\textbf{Adam.} \textbf{Left:} Mean squared error of kernel to normalized arc-cosine kernel and \textbf{Right:} $(E^{n^{(l)}})^2$ as measured against $\varepsilon$. }
\label{fig:phaseadammnist}
\end{figure}

In Figure~\ref{fig:adamkernelsmnist} we show the kernels, samples and learning curves for the networks trained with Adam.

\renewcommand\scalev{0.13}
\renewcommand{\sizespace}{0.13}
\renewcommand{\aesize}{0.8}

\def\directory{nonzero_sum/mnist/adam}

\def\slurmids{77733, 77736, 77739, 77742, 77743, 77745, 77746, 77747}

\def\colorbar{colorbar23000}

\def\figlab{fig:adamkernelsmnist}

\def\figcap{\textbf{Adam.} Inputs with non-zero sums as outlined in \S\ref{sec:exp2}, but with MNIST data. \textbf{Top to bottom:} ${\varepsilon = [10^{-32}, 10^{-16},  10^{-8}, 10^{-6}, 10^{-5}, 10^{-4}, 10^{-2}, 10^{-1}, 1]}$.}

\def\iternum{23000}

\def\filetype{png}
\def\quality{tn}

\ifthenelse{\boolean{figuresbool}}{
\begin{figure}[!htb]
\centering
\foreach \id in \slurmids
{
	\begin{subfigure}[t]{\sizespace \linewidth}
	\includegraphics[scale=\scalev]{figures/\directory/\id/kernel_1_\quality.\filetype} 
	\end{subfigure}
	\begin{subfigure}[t]{\sizespace \linewidth}
	\includegraphics[scale=\scalev]{figures/\directory/\id/kernel_2_\quality.\filetype} 
	\end{subfigure}
	\begin{subfigure}[t]{\sizespace \linewidth}
	\includegraphics[scale=\scalev]{figures/\directory/\id/kernel_3_\quality.\filetype} 
	\end{subfigure}
	\begin{subfigure}[t]{\sizespace \linewidth}
	\includegraphics[scale=\scalev]{figures/\directory/\id/kernel_4_\quality.\filetype} 
	\end{subfigure}
	\begin{subfigure}[t]{\sizespace \linewidth}
	\includegraphics[scale=\scalev]{figures/\directory/\id/kernel_full_\quality.\filetype} 
	\end{subfigure} 
	\begin{subfigure}[t]{\sizespace \linewidth}
	\includegraphics[scale=\scalev]{figures/\directory/\id/metric_loss_\quality.\filetype} 
	\end{subfigure}
	\begin{subfigure}[t]{\sizespace \linewidth}
	\includegraphics[scale=\aesize, trim=-10 -14 0 0]{figures/\directory/\id/ae\iternum.png}
	\end{subfigure} \newline
}
\begin{subfigure}[b]{0.36\textwidth}
\includegraphics[scale=0.3]{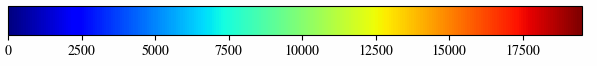}
\end{subfigure}
\caption{\figcap}
\label{\figlab}
\end{figure}
}
{Figures here}

\end{document}